\documentclass{article}
\usepackage{arxiv}

\usepackage[utf8]{inputenc} % allow utf-8 input
\usepackage[T1]{fontenc}    % use 8-bit T1 fonts
\usepackage{hyperref}       % hyperlinks
\usepackage{url}            % simple URL typesetting
\usepackage{booktabs}       % professional-quality tables
\usepackage{amsfonts}       % blackboard math symbols
\usepackage{nicefrac}       % compact symbols for 1/2, etc.
\usepackage{microtype}      % microtypography
\usepackage{lipsum}		% Can be removed after putting your text content
\usepackage{graphicx}
\usepackage{doi}
\usepackage{color}
\usepackage{graphicx}
\usepackage{amsmath}
\usepackage{amssymb}
\usepackage {arydshln}
\usepackage[linesnumbered,algoruled,boxed,lined]{algorithm2e}

\newcommand{\angletheta}{\theta}
\newcommand{\anglephi}{\phi}

\title{ManifoldNeRF: View-dependent Image Feature Supervision \\ for Few-shot Neural Radiance Fields}

\author{ 
        \href{https://orcid.org/0000-0003-2300-4189}{\includegraphics[scale=0.06]{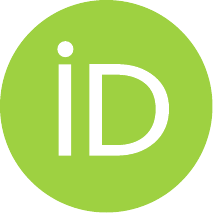}\hspace{1mm}Daiju Kanaoka} \\
	Graduate School of Life Science and Systems Engineering\\
	Kyushu Institute of Technology\\
	Fukuoka, Japan\\
	Guardian Robot Project\\
	RIKEN\\
	Kyoto, Japan \\
	\texttt{kanaoka.daiju327@mail.kyutech.jp} \\
	\And
	\href{https://orcid.org/0000-0001-7429-4011}{\includegraphics[scale=0.06]{orcid.pdf}\hspace{1mm}Motoharu Sonogashira} \\
	Guardian Robot Project\\
	RIKEN\\
	Kyoto, Japan \\
	\texttt{motoharu.sonogashira@riken.jp} \\
	\And
 	\href{https://orcid.org/0000-0002-3669-1371}{\includegraphics[scale=0.06]{orcid.pdf}\hspace{1mm}Hakaru Tamukoh} \\
	Graduate School of Life Science and Systems Engineering\\
	Kyushu Institute of Technology\\
	Fukuoka, Japan\\
        Research Center for Neuromorphic AI Hardware \\
	Kyushu Institute of Technology\\
	Fukuoka, Japan\\
	\texttt{tamukoh@brain.kyutech.ac.jp} \\
	\And
	\href{https://orcid.org/0000-0002-3799-4550}{\includegraphics[scale=0.06]{orcid.pdf}\hspace{1mm}Yasutomo Kawanishi} \\
	Guardian Robot Project\\
	RIKEN\\
	Kyoto, Japan \\
	\texttt{yasutomo.kawanishi@riken.jp} \\
}
\renewcommand{\shorttitle}{ManifoldNeRF}

\hypersetup{
pdftitle={\shorttitle},
pdfsubject={q-bio.NC, q-bio.QM},
pdfauthor={Daiju Kanaoka, Motoharu Sonogashira, Hakaru Tamukoh, Yasutomo Kawanishi},
pdfkeywords={Few-shot NeRF, Manifold}
}

\begin{document}
\maketitle

\begin{abstract}
Novel view synthesis has recently made significant progress with the advent of Neural Radiance Fields (NeRF). DietNeRF is an extension of NeRF that aims to achieve this task from only a few images by introducing a new loss function for unknown viewpoints with no input images. The loss function assumes that a pre-trained feature extractor should output the same feature even if input images are captured at different viewpoints since the images contain the same object. However, while that assumption is ideal, in reality, it is known that as viewpoints continuously change, also feature vectors continuously change. Thus, the assumption can harm training. To avoid this harmful training, we propose ManifoldNeRF, a method for supervising feature vectors at unknown viewpoints using interpolated features from neighboring known viewpoints.  
Since the method provides appropriate supervision for each unknown viewpoint by the interpolated features, the volume representation is learned better than DietNeRF.
Experimental results show that the proposed method performs better than others in a complex scene. We also experimented with several subsets of viewpoints from a set of viewpoints and identified an effective set of viewpoints for real environments. This provided a basic policy of viewpoint patterns for real-world application. The code is available at \url{https://github.com/haganelego/ManifoldNeRF_BMVC2023}
\end{abstract}

% keywords can be removed

\section{Introduction}
\begin{figure*}[tb]
    \centering
    \includegraphics[width=\linewidth]{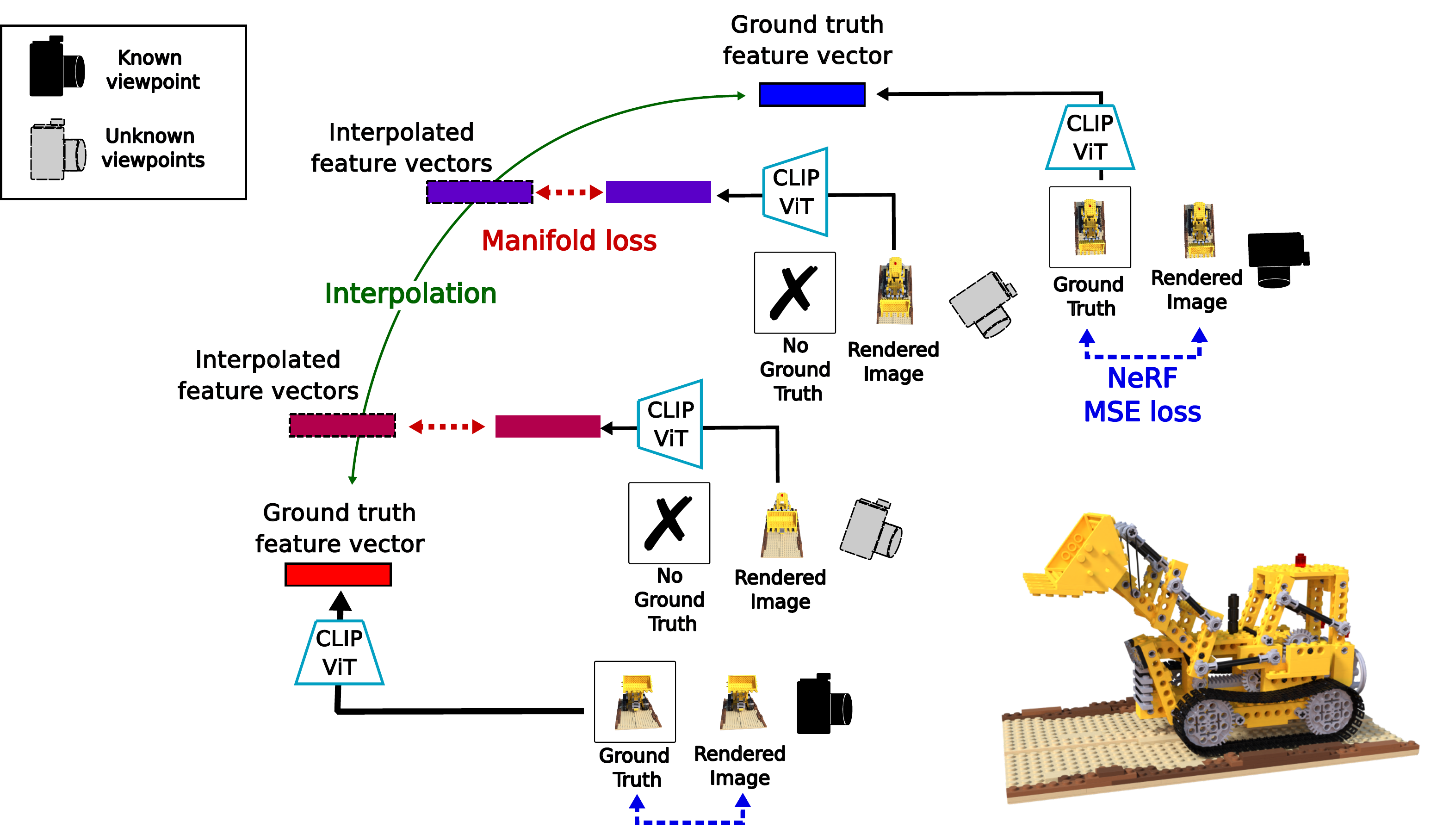}
    \caption{A \textit{vanilla} NeRF is trained by minimizing the mean squared error (MSE) between a rendered image and the ground truth image at each known viewpoint ({\color{blue}NeRF MSE loss}). However, when only a few known viewpoints are available, the volumetric scene function cannot be learned, and the rendering of unknown viewpoints does not perform well. If we could give a loss for the outputs from unknown viewpoints, it can be a clue to learning the volumetric scene function even from a few known viewpoints. We propose a new loss for image features at unknown viewpoints supervised by interpolated features from neighboring known viewpoints ({\color{red}Manifold loss}), which makes it possible to provide supervision even at unknown viewpoints.}
    \label{fig:1_1_overview}
\end{figure*}

Novel view synthesis is one of the challenging tasks to generate an arbitrary viewpoint image from images at a limited number of viewpoints. 
NeRF~\cite{mildenhall2020nerf} is a method that learns a volumetric scene function from images obtained from a scene using an MSE loss and generates arbitrary viewpoint images by volume rendering~\cite{Kajiya1984-aj}. NeRF has been a breakthrough in novel view synthesis, and many methods have been proposed to improve accuracy~\cite{Zhang2020-dw}, training speed~\cite{Muller2022-ws} and application to various fields~\cite{Martin-Brualla2020-tt, Wang2021-wb,zhou2023nerf}.

Although NeRF is a breakthrough method in novel view synthesis, it still has many limitations. One of the most critical limitations is that it requires a lot of viewpoint images.
Recently, the problem of few-shot novel view synthesis has been addressed. DietNeRF~\cite{Jain2021-ru}, one of the few-shot NeRF methods, is based on the assumption that the feature vectors of images of an object at arbitrary viewpoints in the same scene are consistent. In the method, the feature vector at the known viewpoint and feature vectors at arbitrary viewpoints are made closer to each other. Here, the features are obtained from a feature extractor of a pre-trained classifier; we call it ``a pre-trained model''. This enables supervision for any viewpoints that are not included in the training dataset, and it demonstrates high performance in the few-shot novel view synthesis task.

However, are the feature vectors obtained by a pre-trained model actually similar for all viewpoints? In an ideal case, the pre-trained model should output identical feature vectors for images of the same object, but in actual situations, feature vectors vary depending on various factors. Among these factors, viewpoint change makes a drastic change in the observed images, which leads to significant changes in the feature vectors. Also, it has been shown that a continuous change in viewpoint causes a continuous change in the feature vectors in the feature space, as introduced in the Parametric Eigenspace method~\cite{Murase1995-bu}. In the field of pattern recognition, a set of features that varies continuously along some parameters is conventionally called a manifold, and methods such as GAN~\cite{Goodfellow2020-tq} and VAE~\cite{Kingma2013-jx} acquire the manifold as a latent representation, allowing them to generate continuous changes in facial expressions~\cite{Choi2018-zc}, for example.

Based on the above, we reconsider the assumptions used in DietNeRF. While the feature vectors of images from close viewpoints will certainly be similar, the feature vectors of images from very different viewpoints, such as front and back, are likely to be very different. Therefore, DietNeRF's training process may constrain the feature vectors, which are inherently very different, to be close to each other; this harms the training of the volumetric scene function. This can be prevented by providing a relaxed assumption that the feature vectors change as the viewpoint changes.

We propose ManifoldNeRF, a novel few-shot NeRF method based on the concept of the manifold. The overview of ManifoldNeRF is shown in Figure~\ref{fig:1_1_overview}. In this method, we introduce a novel manifold loss for the training of NeRF. For the calculation, a feature vector at an arbitrary viewpoint is interpolated by the neighboring known viewpoints. The loss is calculated from the interpolated feature vector, named a pseudo ground truth. The difference between the feature vectors extracted from the rendered images at arbitrary viewpoints and the pseudo ground truth is used as the auxiliary loss, as well as the original MSE loss, to train a model. Our contributions are as follows:
\begin{itemize}
    \item We propose ManifoldNeRF for the few-shot novel view synthesis. We introduce a novel loss function named manifold loss to give supervision for arbitrary viewpoints based on the concept of the Parametric Eigenspace~\cite{Murase1995-bu}.
    \item We show that by interpolating the feature vectors  of neighboring viewpoints, a feature vector that is a good approximation of the ground truth of an interpolated viewpoint can be generated. This enables us to prepare a pseudo ground truth of unknown viewpoints and enables training possible even when the number of known viewpoints is limited.
    \item We conducted experiments to clarify which viewpoints are important for training a volumetric scene function with a limited number of images. These results establish a basic policy for capturing images for real-world applications.
\end{itemize}

\section{Related works}
\subsection{Few-shot NeRF}
NeRF~\cite{mildenhall2020nerf} is a method that embeds a volumetric scene into a volumetric scene function implemented by a multilayer perceptron (MLP), which output color $\mathbf{c} = (r,g,b)^\top$ and volume density $\sigma$, 3D location $\mathbf{x} = (x,y,z)^\top$ and direction $\mathbf{d} = (\angletheta, \anglephi)^\top$ as input to MLP.
The value of each pixel in an image is calculated by differentiable rendering that enables backpropagation.
The loss function is the MSE loss which is commonly used in image reconstruction tasks such as an autoencoder~\cite{hinton2006-ae}. 

One drawback of \textit{vanilla} NeRF is that it requires a lot of images for training. To address this problem, various approaches have been proposed, such as using feature vectors obtained from pre-training models~\cite{Jain2021-ru, Yu2021-qx,chen2021mvsnerf, wang2021ibrnet}, optimization for depth using sparse 3D points~\cite{deng2022depth} and reducing artifact generation by constraining each point on the ray~\cite{kim2022infonerf, niemeyer2022regnerf}.

\textbf{DietNeRF:}  
In addition to the MSE loss of the \textit{vanilla} NeRF, DietNeRF introduces a semantic consistency loss to enable training with few images. The loss provides a constraint to force the feature vector obtained from an image at a randomly sampled known viewpoint and the feature vector of a rendered image at an arbitrary viewpoint to be closer.
In the method, a pre-trained vision transformer~\cite{Dosovitskiy2020-uv} of CLIP~\cite{Radford2021-cm}, which we call it ``CLIP-ViT'', is used as a feature extractor.
Equation~\ref{eq:Lsc} show the semantic consistency loss $\mathcal{L}_{\text{SC}}$ between feature vectors $\mathbf{v}_k, \mathbf{v}_u$ of known and unknown viewpoint is measured by cosine similarity, where $\lambda$ is a scaling factor.
\begin{equation}
    \mathcal{L}_{\text{SC}}(\mathbf{v}_k, \mathbf{v}_u) = \lambda ( 1 -\mathbf{v}_k^\top\mathbf{v}_u ),
    \label{eq:Lsc}
\end{equation}

Experiments conducted on DietNeRF have shown that the semantic consistency loss is effective when the number of pixels selected is 15\%$\sim$20\% of the total number of pixels.
This method can be trained from a one-shot image and is currently one of the most effective few-shot NeRF methods.

In the proposed method, we select an unknown viewpoint from the neighborhood of the known viewpoints, and we interpolate the feature vector of a viewpoint from the feature vectors of the known viewpoints to obtain pseudo ground truth, aiming to give richer supervision than DietNeRF.

\subsection{Manifold}
\label{2_Manifold}

% Parametric Eigenspaceの話
When a camera pose relative to an object changes smoothly, the observed image will also change smoothly. Hence, we can assume that the feature vector extracted from the observed image also changes smoothly according to the camera motion.
Based on this assumption, the Parametric Eigenspace method, proposed by Murase et al.~\cite{Murase1995-bu}, interpolates feature vectors at intermediate camera views from two adjacent existing camera views.
On the basis of the key concept, the Parametric Eigenspace can model a 3D object using a few images.
The method uses a low-dimensional eigenspace calculated by the principal component analysis~(PCA) as a feature space and interpolates intermediate image features in the feature space.
% Manifoldとは何かの話（近傍の点間のみで演算が意味をなすところから，慣例的に多様体と呼ばれている）局所的にユークリッド空間とみなせる
Here, since a camera pose has six degrees of freedom, the feature vectors are mapped on a 6D hyperplane in the Eigenspace according to the camera parameter.
On the hyperplane, feature vectors at similar viewpoints are mapped onto similar points and can be used for interpolation, while calculation using two points far from each other is meaningless.
Traditionally, this hyperplane is called a \textit{manifold} in this field.

% Deep Manifold Embeddingの話
Ninomiya et al.~\cite{Ninomiya2017-vp} have extended the concept of the Parametric Eigenspace; they propose the deep manifold embedding that finds a low-dimensional feature space using deep Learning.
The method can find more suitable feature spaces for object pose estimation than PCA.
% Ω-GANの話
Kawanishi et al.~\cite{Kawanishi2021-ga} have proposed to use this concept for the latent space of GANs.
The method samples a latent variable from a distribution on the manifold corresponding to the camera pose; it makes GANs able to generate an image from a specific viewpoint intuitively.
% 最近はいろいろなところに使われている\cite{Choi2018-zc, Gong2022-nk, Zang2022-sm}
The concept, the Deep Manifold Embedding, has been used in various applications~\cite{Choi2018-zc, Gong2022-nk, Zang2022-sm} to find a feature space that can model the target with variations.

In this study, we borrow the concept from the Parametric Eigenspace and the Deep Manifold Embedding; we aim to interpolate feature vectors at unknown viewpoints from feature vectors at adjacent known viewpoints to calculate an additional loss for the NeRF training.

\section{Feature vector difference according to viewpoints}
In this section, we show the results of preliminary experiments that investigate the relationship between viewpoints and feature vectors obtained from a pre-trained model. We used images of the LEGO scene from the NeRF synthetic dataset~\cite{mildenhall2020nerf}.

\begin{figure}[tb]
    \centering
    \includegraphics[width=0.8\linewidth]{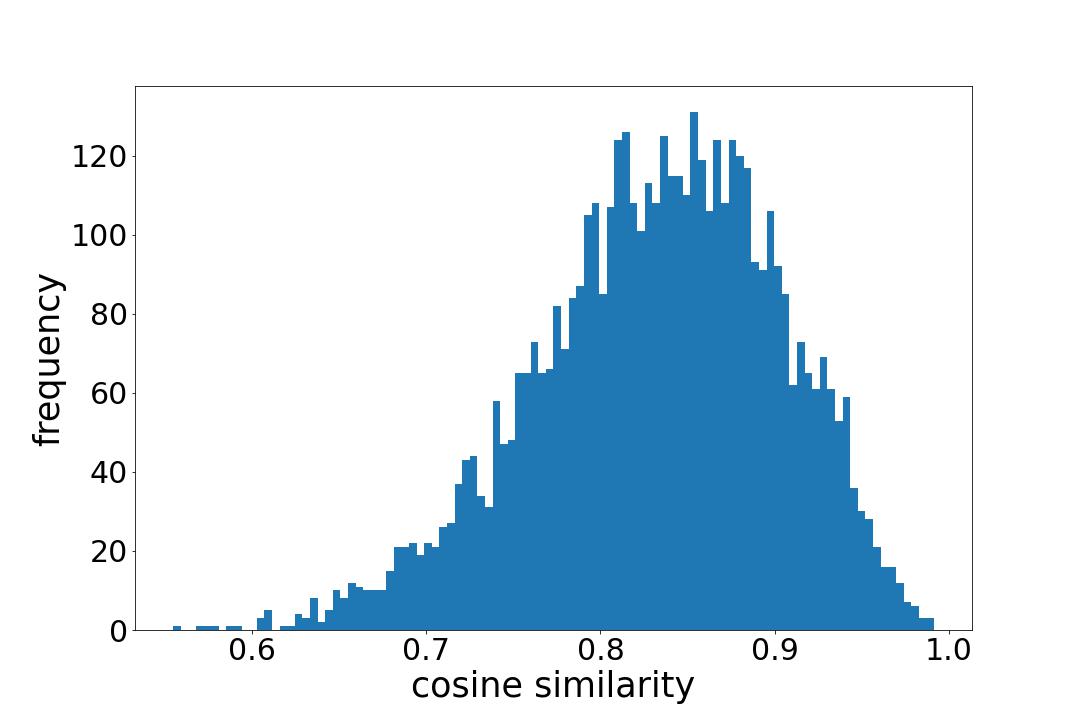}
    \caption{Histogram of cosine similarity scores for all pairs of the 100 images in the LEGO training dataset}
    \label{fig:3_1_cos_sim_hist}
\end{figure}

First, Figure~\ref{fig:3_1_cos_sim_hist} shows a histogram of the cosine similarity scores for all pairs of the 100 images in the training dataset. The lowest cosine similarity score was about 0.5. We presume that pairs with significantly different viewpoints are included in the area where the cosine similarity score is low.

\begin{figure}[tb]
    \begin{minipage}[b]{0.48\columnwidth}
        \centering
        \includegraphics[width=\linewidth]{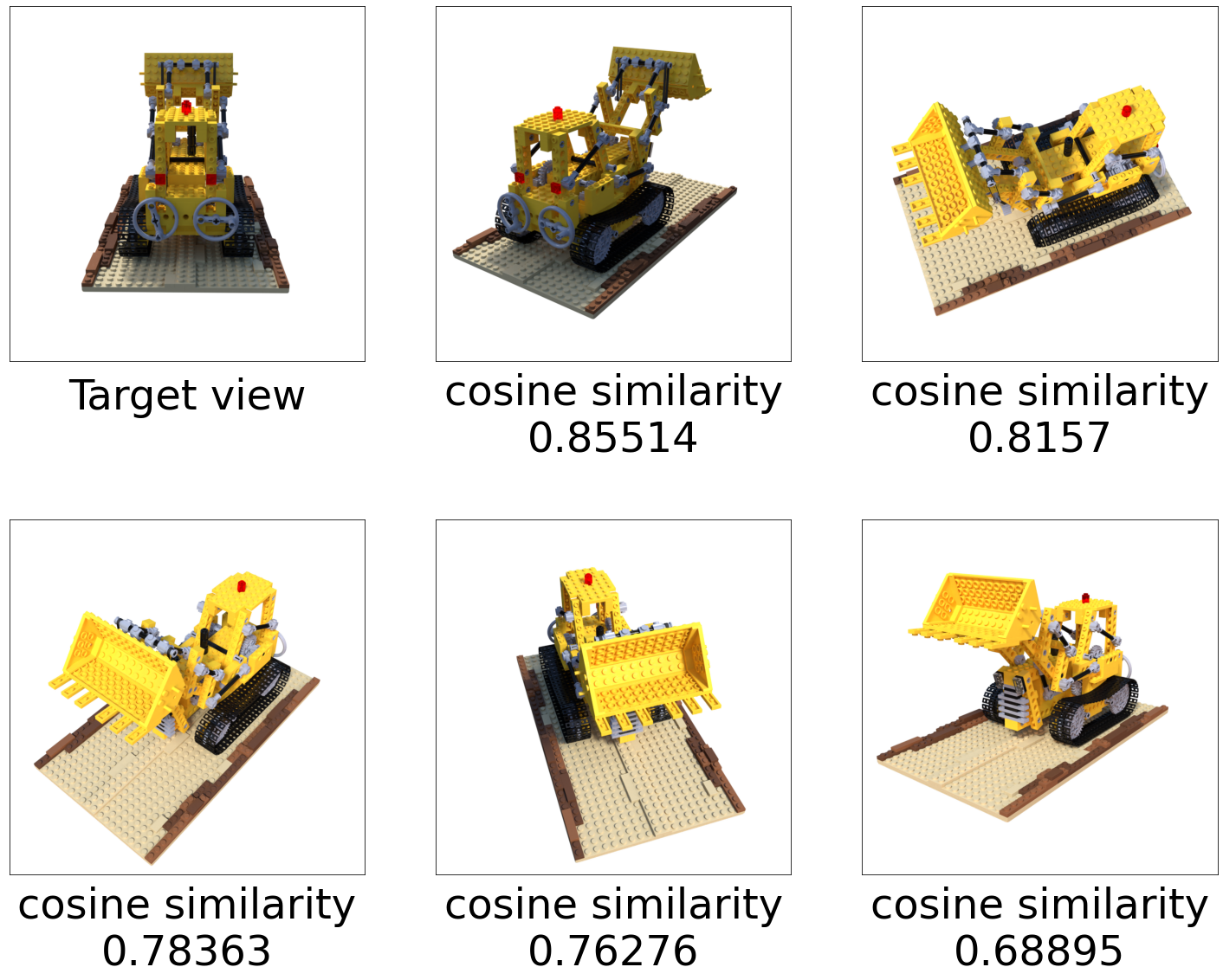}
        \caption{The cosine similarity scores are relatively high for viewpoints that are close to the target view, but the cosine similarity scores are relatively low for viewpoints that are largely far from the target view.}
        \label{fig:3_2_comparison_with_targetview}
    \end{minipage}
    \hspace{0.01\columnwidth}
    \begin{minipage}[b]{0.48\columnwidth}
        \centering
        \includegraphics[width=\linewidth]{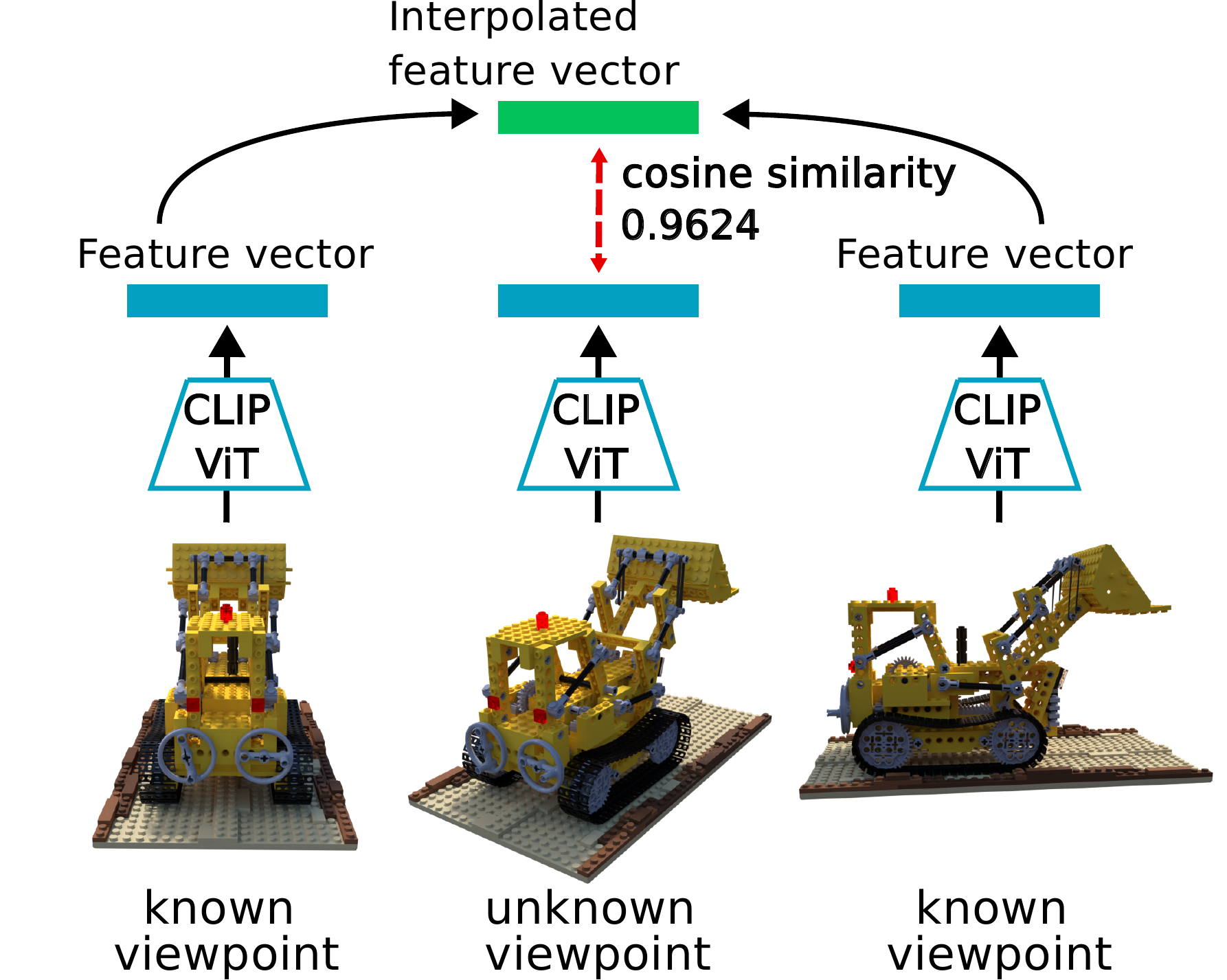}
        \caption{The interpolated feature vectors had a higher cosine similarity score than the cosine similarity scores of the feature vectors obtained from CLIP-ViT for different viewpoints.}
        \label{fig:3_3_comparison_with_interp}
    \end{minipage}

\end{figure}

Figure~\ref{fig:3_2_comparison_with_targetview} shows the comparison of the cosine similarity between the image from the back (target view) of the LEGO and other viewpoints in the training dataset.
The cosine similarity score was relatively high when the viewpoint was close to the target view; in contrast, the cosine similarity score was relatively low when the viewpoint differed significantly from the target view.
From the above, we expect that the loss function given by DietNeRF would not be so effective when the viewpoints are largely different since it violates the assumption of DietNeRF, while it works when the viewpoints are close to each other.

We further investigated the interpolation of the feature vectors at an unknown viewpoint from two neighboring known viewpoints.
The back and side of the LEGO in the training dataset are used as the images at known viewpoints, and a viewpoint between them is selected as an unknown viewpoint.
We compared the feature vectors at the three known/unknown viewpoints using cosine similarity.
We also compared the groud-truth feature vector at the unknown viewpoint with an interpolated feature from two known viewpoints, which is simply calculated by averaging the two vectors at the known viewpoints, assuming that the unknown viewpoint is exactly in the middle of the two known viewpoints.

The experimental results are shown in Figure~\ref{fig:3_3_comparison_with_interp}. The cosine similarity score with the interpolated feature vector was 0.9624, indicating that the interpolated feature vector was closer to the ground-truth feature vector of the image at the unknown viewpoint.

\section{Proposed method: ManifoldNeRF}
Unlike DietNeRF, the proposed method requires an unknown viewpoint to be located between two known viewpoints. Since most 360-degree view scenes are taken from cameras located in a hemisphere from the center of the scene, an arbitrary viewpoint image is generated based on this assumption.
Thus, we select an unknown viewpoint $\mathbf{p}_{u}$ by interpolating two known viewpoints $\mathbf{p}_{k,1}$ and $\mathbf{p}_{k,2}$ using spherical linear interpolation (Slerp)~\cite{Shoemake1985-yr};
%When generating a viewpoint between two known viewpoints $\mathbf{p}_{k,1}, \mathbf{p}_{k,2}$ spherical linear interpolation (Slerp)~\cite{Shoemake1985-yr} is used to generate an unknown viewpoint $\mathbf{p}_{u}$:
\begin{align}
    \mathbf{p}_{u} &= \frac{\sin{(1-s)\theta}}{\sin{\theta}}\mathbf{p}_{k,1} + \frac{\sin{s}\theta}{\sin{\theta}}\mathbf{p}_{k,2},\\
    \theta &= \arccos{(\mathbf{p}_{k,1}, \mathbf{p}_{k,2})},
\end{align}
where $s$ is an interpolation coefficient, which is a random value following a uniform distribution in the range [0,~1].
Each pair of two known viewpoints for choosing unknown viewpoints cannot be too distant since the computation is meaningful only between close points on a manifold; interpolation of feature vectors obtained from the pre-trained model cannot work well between viewpoints that are too distant.
%If a pair of two viewpoints cannot be set up in advance, the pair is selected if the distance between the two viewpoints is less than a threshold.
Pairs of known viewpoints are sampled if the distance between them is less than a threshold.
When interpolating a feature vector $\widehat{\mathbf{v}}_u$ between feature vectors $\mathbf{v}_{k,1}, \mathbf{v}_{k,2}$ of two viewpoints $\mathbf{p}_{k,1}$ and $\mathbf{p}_{k,2}$ is performed by linear interpolation (Lerp) using the following formula:
\begin{equation}
    \widehat{\mathbf{v}}_u = (1-s)\mathbf{v}_{k,1} + s\mathbf{v}_{k,2}.
\end{equation}
The training process is the same as that of DietNeRF except that we replace the semantic consistency loss with the manifold loss. The manifold loss is the cosine similarity between the feature vector $\mathbf{v}_u$ of a rendered image at an arbitrary viewpoint and the feature vector $\widehat{\mathbf{v}}_u$ interpolated from known neighboring viewpoints and is calculated by
\begin{equation}
\mathcal{L}_{\text{ML}}(\mathbf{v}_u, \widehat{\mathbf{v}}_u) = \lambda(1 - \mathbf{v}_u^\top\widehat{\mathbf{v}}_u).
\end{equation}

\section{Experiments}

In this study, all experiments used a 40GB NVIDIA A100. The number of training iterations was 100,000.
%In Experiments~\ref{4_ex1} and~\ref{experiment: uniform}, pairs of neighboring viewpoints could not be set in advance, so pairs were set as those with a distance between neighboring viewpoints of at least 2.5.
In Experiments~\ref{4_ex1}, ~\ref{experiment: uniform} and ~\ref{experiment:dtu}, viewpoint pairs were selected using a threshold of 2.5, while viewpoint pairs were predefined in Experiment~\ref{4_ex3}.
We used PSNR, SSIM~\cite{Wang2004-sc}, and LPIPS~\cite{Zhang2018-fv} as evaluation metrics. 
We used the following two public datasets to evaluate the proposed method.

\textbf{NeRF synthetic dataset:} This dataset contains 100 images in each of the 8 scenes by cameras randomly placed in a hemispherical pattern from the center of each object~\cite{mildenhall2020nerf}. To evaluate the performance of the proposed method in few-shot scenarios, we randomly selected 8 images as the training dataset for each scene.   

\textbf{DTU MVS dataset:} This dataset captures physical object in real environmets and contains 49 images in each of the 128 scenes~\cite{DTU}. We conducted experiments on the 8 scenes and randomly selected 8 images as the training dataset while testing with the remaining 41 images.

\subsection{Randomly sampled NeRF synthetic dataset}
\label{4_ex1}
Table~\ref{tab:per_scene_realistic_synth} shows the experimental results for each scene. ManifoldNeRF was competitive with DietNeRF and InfoNeRF depending on the scenes, and we could not conclude which method was better in overall performance from this experiment alone.

One of the reasons why the proposed method sometimes performed not well is that the effectiveness of the manifold loss strongly depends on the locations of the known viewpoints.
In DietNeRF, the loss can be calculated for any viewpoint in the training process due to the assumption that all feature vectors of arbitrary viewpoints in the same scene should be the same.
In contrast, the proposed method can only compute the loss for viewpoints between two neighboring known viewpoints closer than a threshold. Therefore, the proposed method may not perform well depending on the bias of the viewpoints in the training dataset.
\begin{table*}
\caption{The results of training 8 randomly selected images in each scene of the NeRF synthetic dataset. The highest score is in \textbf{bold}, and the second highest score is \underline{underlined}.}
\label{tab:per_scene_realistic_synth}
\centering
\setlength{\tabcolsep}{3.65pt}
\small 
\begin{tabular}{@{}lcccccccc@{}}
\toprule
\textbf{PSNR} $\uparrow$  & \textbf{LEGO} & \textbf{Chair} & \textbf{Drums} & \textbf{Ficus} & \textbf{Mic} & \textbf{Ship} & \textbf{Materials} & \textbf{Hotdog} \\ \midrule
NeRF & 9.727 & 18.920 & 17.032 & 19.894 & 12.889 & 19.535 & 7.945 & 10.561\\
InfoNeRF & 9.667 & \underline{24.964} & \textbf{19.116} & \underline{20.924} & 24.233 & 20.299 & 20.773 & 10.752 \\
DietNeRF  & \underline{22.063} & 24.722 & 18.757 & \textbf{21.173} & \textbf{26.383} & \underline{22.480} & \underline{20.671} & \underline{25.746} \\
ManifoldNeRF (ours) & \textbf{22.171} & \textbf{26.155} & \underline{18.900} & 20.489 & \underline{25.967} & \textbf{23.001} & \textbf{20.819} & \textbf{26.321} \\\midrule
NeRF, 100 views & \textbf{30.336} & \textbf{32.807} & \textbf{25.144} & \textbf{28.174} & \textbf{32.196} & \textbf{28.654} & \textbf{28.459} & \textbf{35.758}\\\bottomrule
\end{tabular}
\end{table*}

\subsection{Uniformly sampled NeRF synthetic dataset}
\label{experiment: uniform}

\begin{figure}[tb]
    \begin{minipage}[b]{0.12\columnwidth}
        \centering
        \includegraphics[width=\columnwidth]{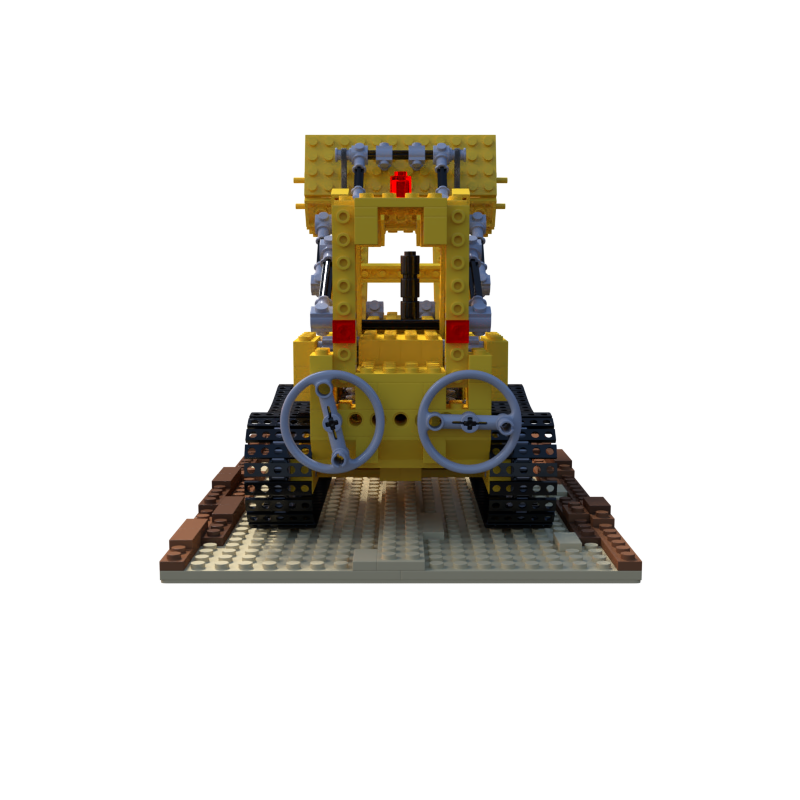}
    \end{minipage}
    \begin{minipage}[b]{0.12\columnwidth}
        \centering
        \includegraphics[width=\columnwidth]{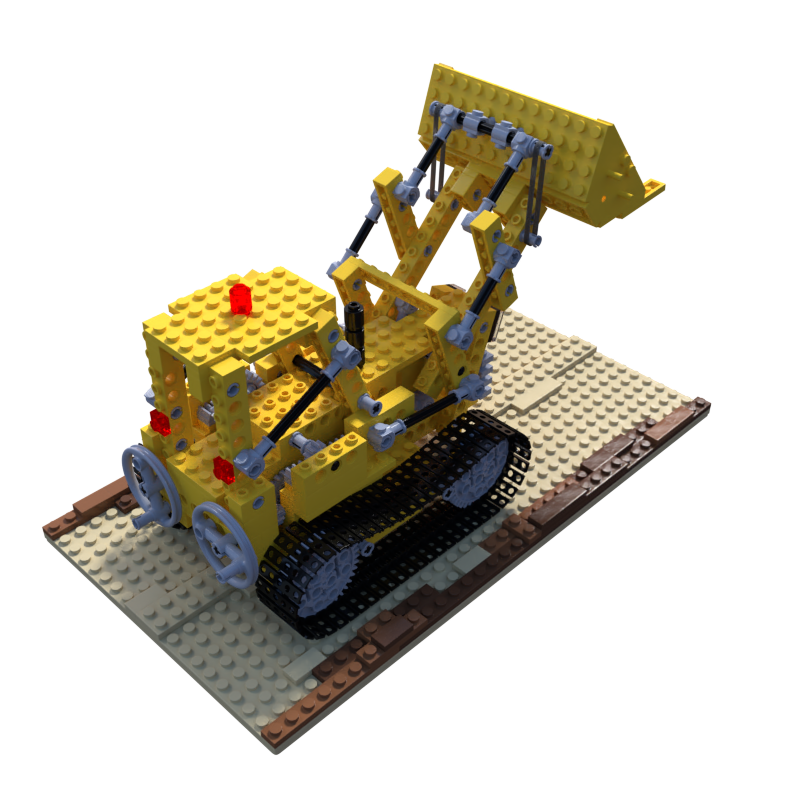}
    \end{minipage}
    \begin{minipage}[b]{0.12\columnwidth}
        \centering
        \includegraphics[width=\columnwidth]{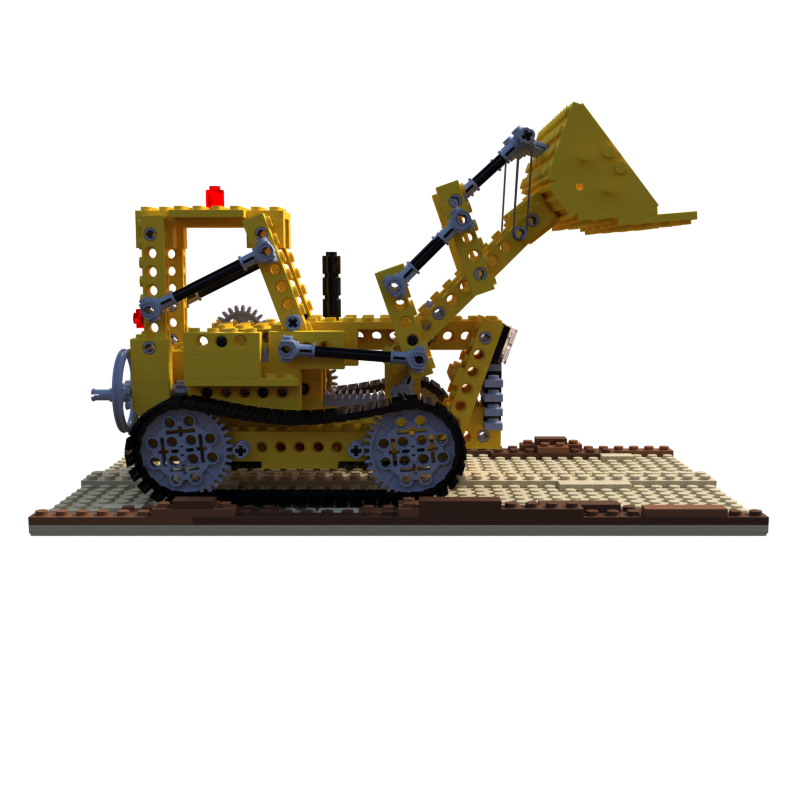}
    \end{minipage}
    \begin{minipage}[b]{0.12\columnwidth}
        \centering
        \includegraphics[width=\columnwidth]{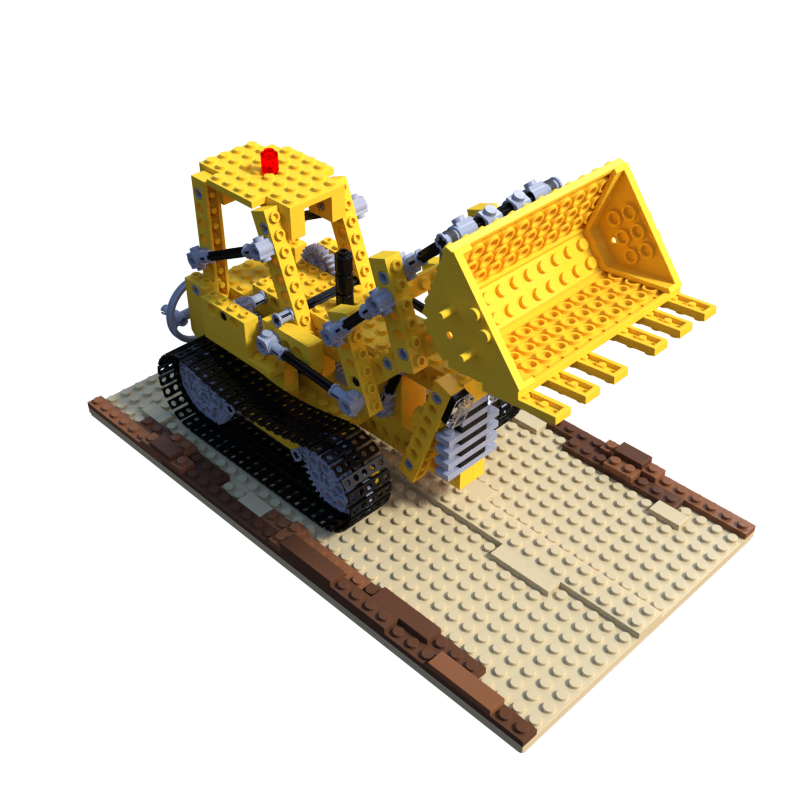}
    \end{minipage}
    \begin{minipage}[b]{0.12\columnwidth}
        \centering
        \includegraphics[width=\columnwidth]{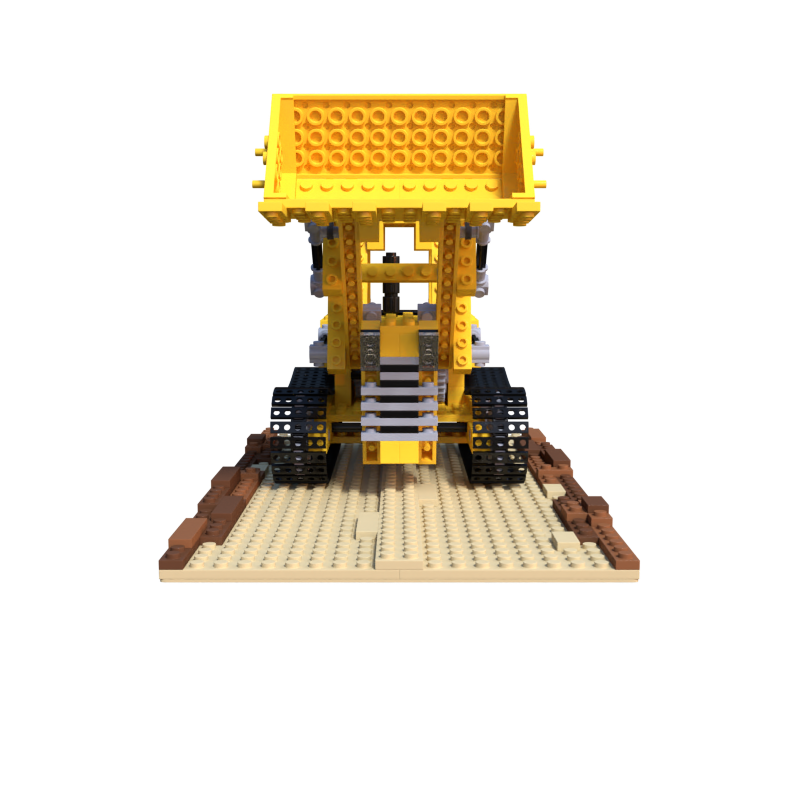}
    \end{minipage}
    \begin{minipage}[b]{0.12\columnwidth}
        \centering
        \includegraphics[width=\columnwidth]{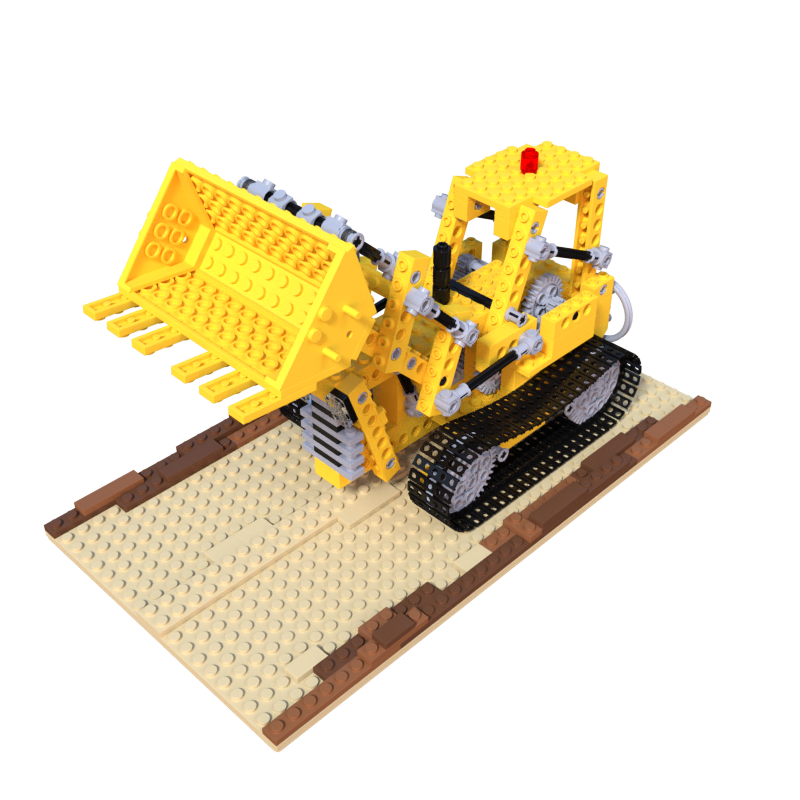}
    \end{minipage}
    \begin{minipage}[b]{0.12\columnwidth}
        \centering
        \includegraphics[width=\columnwidth]{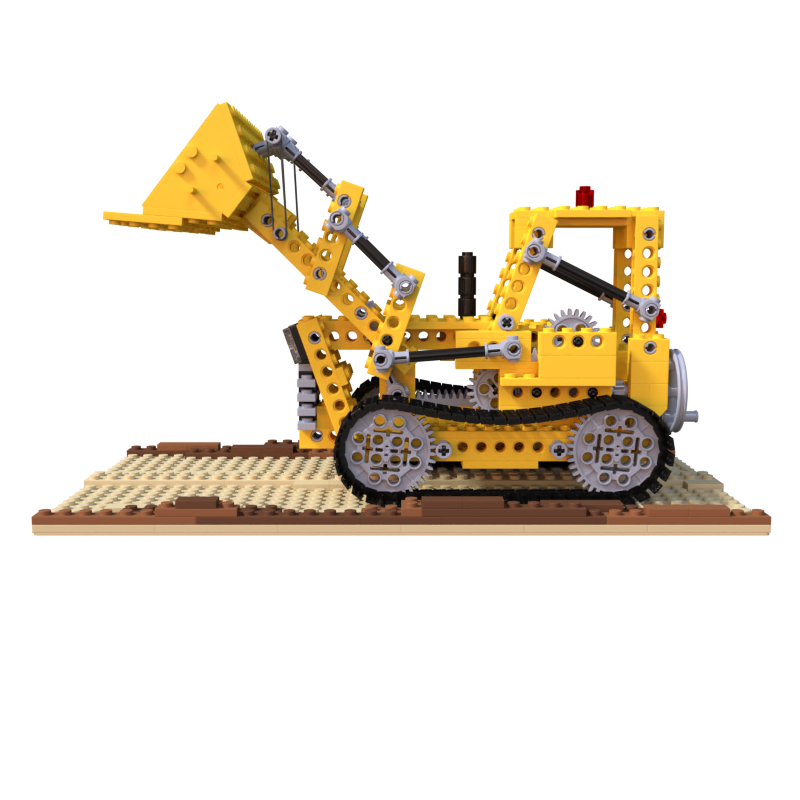}
    \end{minipage}
    \begin{minipage}[b]{0.12\columnwidth}
        \centering
        \includegraphics[width=\columnwidth]{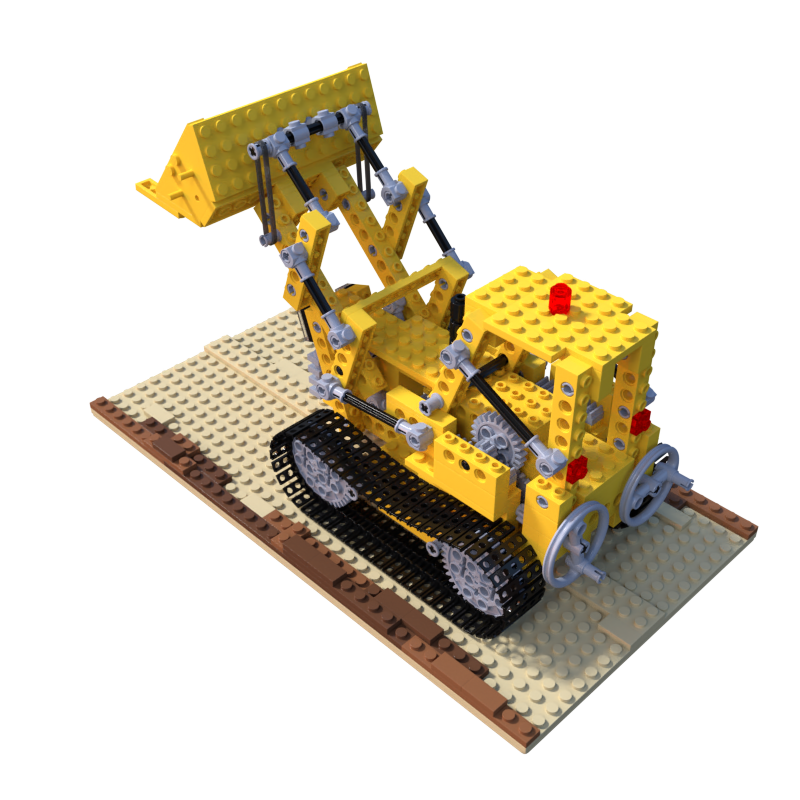}
    \end{minipage}
    \caption{Viewpoint images of the LEGO training dataset for the experiment in Sec.~\ref{experiment: uniform}.}
    \label{fig:uniformed_lego_dataset}
\end{figure}

\begin{figure}[tb]
    \centering
    \begin{minipage}[b]{0.25\columnwidth}
        \centering
        \includegraphics[width=\columnwidth]{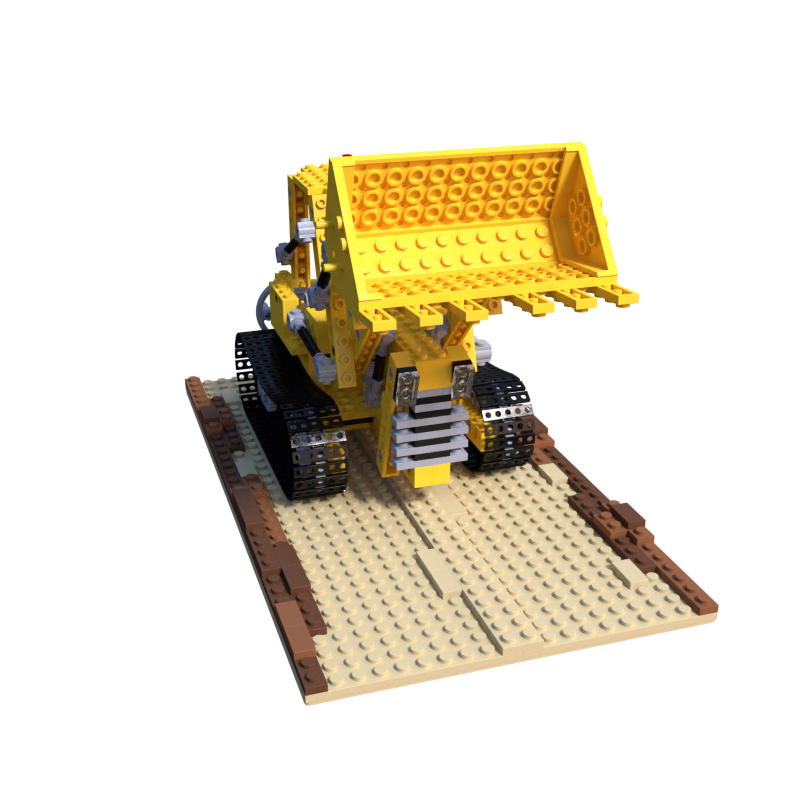}
        Ground Truth
    \end{minipage}
    \hspace{0.01\columnwidth}
    \begin{minipage}[b]{0.25\columnwidth}
        \centering
        \includegraphics[width=\columnwidth]{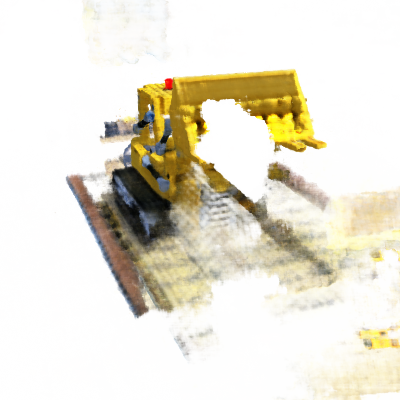}
        DietNeRF
    \end{minipage}
    \hspace{0.01\columnwidth}
    \begin{minipage}[b]{0.25\columnwidth}
        \centering
        \includegraphics[width=\columnwidth]{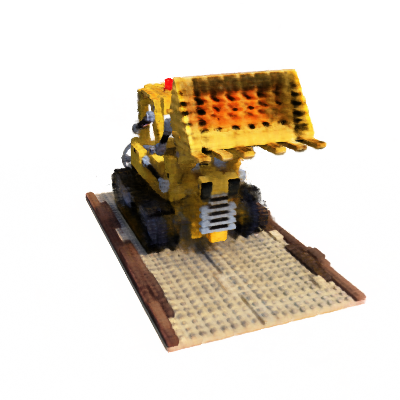}
        ManifoldNeRF(ours)
    \end{minipage}
    \label{fig:uniformed_lego}
    \caption{Qualitative comparison on the NeRF synthetic dataset for the experiment in Sec.~\ref{experiment: uniform}}
\end{figure}

Based on the experiments in the previous section, we selected images for training and evaluation to avoid bias. This experiment was conducted in the LEGO scene of NeRF synthetic dataset. We generated 8 viewpoint images from the Blender model that were evenly distributed across the scene, with 4 images for each side and 4 images for the overhead view as shown in Figure~\ref{fig:uniformed_lego_dataset}. The experimental results are shown in Table~\ref{tab:hardcoding_result} and Figure~\ref{fig:uniformed_lego}. These results show that the proposed method achieves better PSNR+7.471, SSIM+0.146, and LPIPS-0.150 than DietNeRF, confirming the superior performance of the proposed method. 

In this experiment, the performance of DietNeRF was low compared to the experiment of Sec.~\ref{4_ex1}. 
We consider that this is due to constraints on DietNeRF. As shown in Figure~\ref{fig:3_1_cos_sim_hist}, the cosine similarity is varied, and Figure~\ref{fig:3_2_comparison_with_targetview} shows that the lowest cosine similarity is 0.68. From the above, the constraints on DietNeRF may bring feature vectors that are very different close together. On the other hand, the proposed method showed high performance when the viewpoints were evenly distributed across the scene. Considering real-world applications, we can control the viewpoints to be taken. Therefore, the viewpoint constraint in the proposed method is not a drawback in real-world applications.

\begin{table}[tb]
\centering
\caption{The result of training using images at uniformly selected viewpoints from the LEGO scene of the NeRF synthetic dataset. The highest score is in \textbf{bold}.}
\label{tab:hardcoding_result}
\begin{tabular}{@{}lccccc@{}}
\toprule
\textbf{Method}         & \textbf{PSNR} $\uparrow$ & \textbf{SSIM} $\uparrow$ & \textbf{LPIPS} $\downarrow$ \\ \midrule
DietNeRF & 15.233 & 0.713 & 0.267  \\
ManifoldNeRF (ours) & \textbf{22.704} & \textbf{0.859} & \textbf{0.117}  \\\bottomrule
\end{tabular}
\end{table}

\subsection{Randomly sampled DTU MVS dataset}
\label{experiment:dtu}
\begin{figure}[tb]
\centering
    \begin{minipage}[t]{0.18\columnwidth}
        \centering
        \includegraphics[width=\columnwidth]{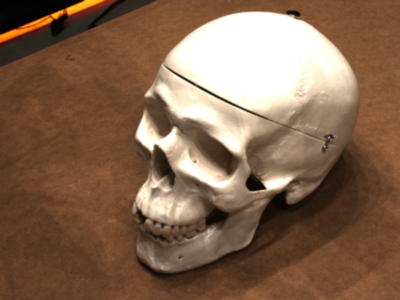}
        \includegraphics[width=\columnwidth]{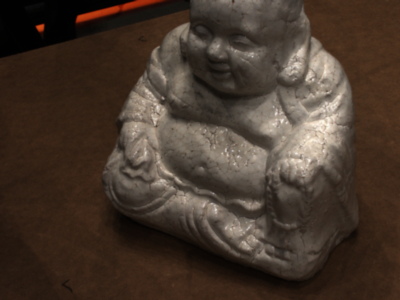}
        Ground Truth  
    \end{minipage}
    \begin{minipage}[t]{0.18\columnwidth}
        \centering
        \includegraphics[width=\columnwidth]{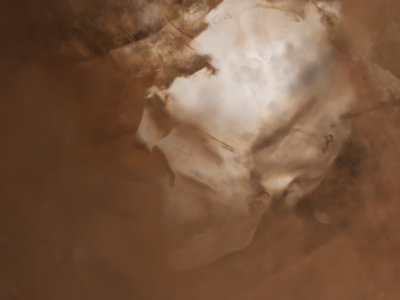}
        \includegraphics[width=\columnwidth]{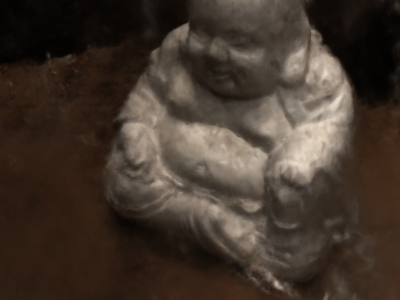}
        NeRF  
    \end{minipage}
    \begin{minipage}[t]{0.18\columnwidth}
        \centering
        \includegraphics[width=\columnwidth]{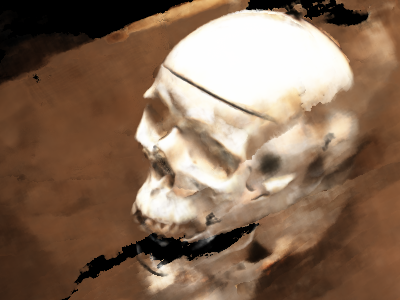}
        \includegraphics[width=\columnwidth]{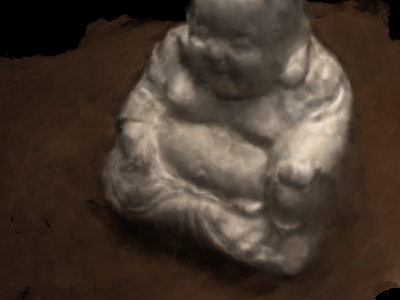}
        InfoNeRF  
    \end{minipage}
    \begin{minipage}[t]{0.18\columnwidth}
        \centering
        \includegraphics[width=\columnwidth]{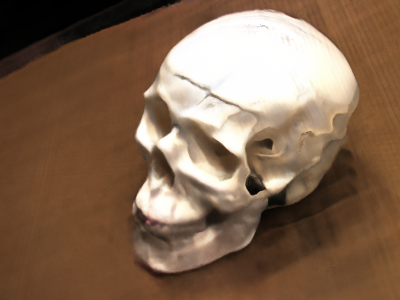}
        \includegraphics[width=\columnwidth]{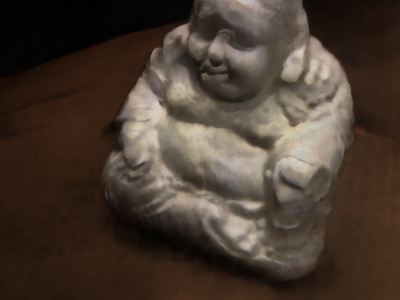}
        DietNeRF
    \end{minipage}
    \begin{minipage}[t]{0.18\columnwidth}
        \centering
        \includegraphics[width=\columnwidth]{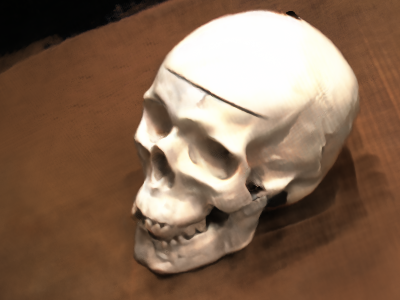}
        \includegraphics[width=\columnwidth]{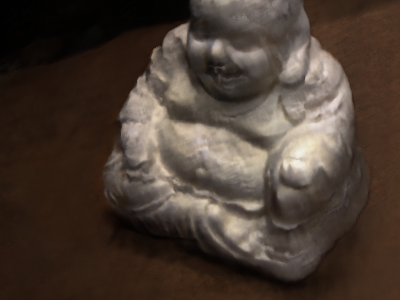}
        ManifoldNeRF (ours)
    \end{minipage}
    \caption{Qualitative comparison on the \#65(\textbf{Top}) and \#114(\textbf{Bottom}) of the MVS DTU dataset for the experiment in Sec~\ref{experiment:dtu}}
    \label{fig:dtu}
\end{figure}

Table~\ref{tab:dtu_result} and Figure~\ref{fig:dtu} shows the experimental result for \#65 and \#114 scenes. 
We confirmed that the proposed method achieved the best performance in this experiment with two scenes.

We consider that the high performance of the proposed method in this experiment is due to its constraint, which can be optimized well in complex scenes such as the DTU MVS dataset.
The NeRF synthetic dataset is generated from a 3D model in Blender, so there is less noise in the image. On the other hand, datasets taken in real environments are likely to contain disturbance due to background effects and other factors. Therefore, the feature vectors at different viewpoints in the same scene are likely to differ significantly. The constraints of the proposed method generate feature vectors between neighboring viewpoints considering disturbances. From the above, we believe that the proposed method performs well in real environments.

\begin{table}[tb]
\caption{The results of training 8 randomly selected images in \#65 and \#114 scene of the DTU MVS dataset. The highest score is in bold, and the second-highest score is underlined.}
\label{tab:dtu_result}
    \begin{minipage}[b]{0.48\columnwidth}
        \centering
        \scalebox{0.75}{
            \begin{tabular}{@{}lccccc@{}}
            \toprule
            \textbf{\#65} & \textbf{PSNR} $\uparrow$ & \textbf{SSIM} $\uparrow$ & \textbf{LPIPS} $\downarrow$ \\ \midrule
            NeRF & 11.970 & 0.481 & 0.527 \\
            InfoNeRF & 14.786 & 0.484 & 0.431  \\
            DietNeRF & \underline{20.883} & \underline{0.698} & \underline{0.352}  \\
            ManifoldNeRF (ours) & \textbf{22.197} & \textbf{0.702} & \textbf{0.302}  \\\bottomrule
            \end{tabular}
        }
    \end{minipage}
    \hspace{0.01\columnwidth}
    \begin{minipage}[b]{0.48\columnwidth}
        \centering
        \scalebox{0.75}{
            \begin{tabular}{@{}lccccc@{}}
            \toprule
            \textbf{\#114}         & \textbf{PSNR} $\uparrow$ & \textbf{SSIM} $\uparrow$ & \textbf{LPIPS} $\downarrow$ \\ \midrule
            NeRF & 18.691 & 0.636 & 0.396  \\
            InfoNeRF & \underline{21.382} & 0.611 & 0.364  \\
            DietNeRF & 20.861 & \underline{0.673} & \underline{0.337}  \\
            ManifoldNeRF (ours) & \textbf{23.202} & \textbf{0.732} & \textbf{0.299}  \\\bottomrule
            \end{tabular}
        }
    \end{minipage}
\end{table}

\subsection{Evaluation of viewpoint patterns for real-world application}
\label{4_ex3}
\begin{figure}[tb]
    \centering
    \includegraphics[width=\linewidth]{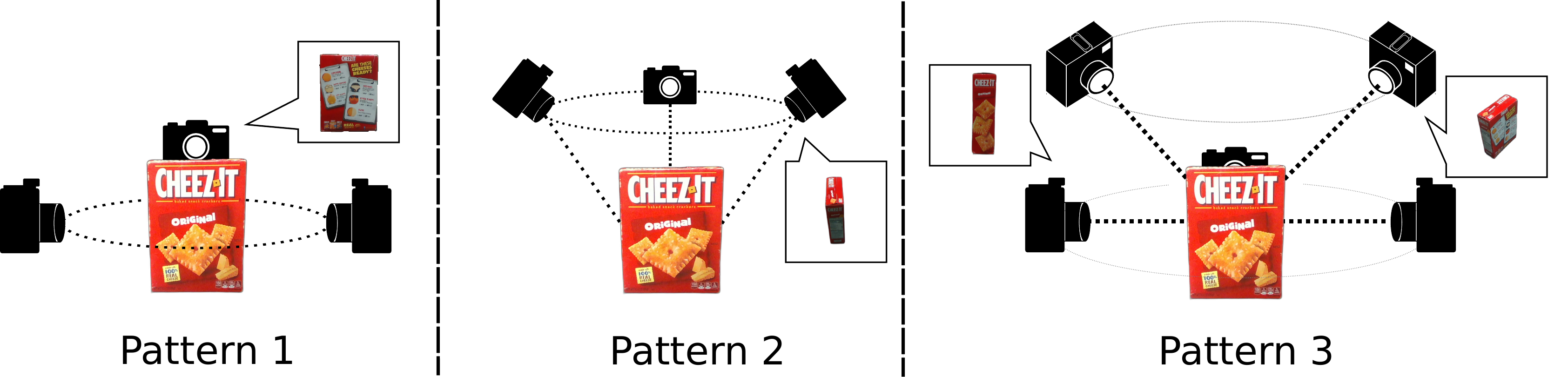}
    \caption{Patterns for taking images for Experiment~\ref{4_ex3}}
    \label{fig:5_2_cheezit_patern}
\end{figure}

In this section, we evaluate several viewpoint patterns to see which viewpoints are appropriate for training ManifoldNeRF from images taken in real environments. In this experiment, we used ``cheezit'', one of the Yale-CMU-Berkeley object~\cite{Calli2015-zu}, which is used as a common benchmark dataset for object manipulation and object recognition. This object was taken images using a Realsense D435, and the image was center-cropped and background-removed. As shown in Figure~\ref{fig:5_2_cheezit_patern}, each viewpoint pattern was taken from 8 directions around the object: (1)~horizontally toward the object, (2)~diagonally above the object, and (3)~alternate horizontally and diagonally to the object.

\begin{table}[tb]
\centering
\setlength{\tabcolsep}{5pt}
\caption{Experimental results for each viewpoint pattern. The highest score is in \textbf{bold}, and the second highest score is \underline{underlined}.}
\label{tab:result_of_cheezit}
\begin{tabular}{@{}lccccc@{}}
\toprule
\textbf{ManifoldNeRF}  & \textbf{PSNR} $\uparrow$ & \textbf{SSIM} $\uparrow$ & \textbf{LPIPS} $\downarrow$ \\ \midrule
Pattern1 & 17.983 & 0.849 & 0.147  \\
Pattern2 & \underline{21.143} & \underline{0.871} & \underline{0.114}  \\
Pattern3 & \textbf{23.203} & \textbf{0.899} & \textbf{0.074}  \\\bottomrule
\end{tabular}
\end{table}

Table~\ref{tab:result_of_cheezit} shows the results of training on each pattern with the proposed method. The proposed method had the best performance for Pattern 3.

The reason why viewpoint Pattern 3 had the best performance in the proposed method is due to the wide range of unknown viewpoints for which feature vector interpolation was possible. In Patterns 1 and 2, we could only interpolate feature vectors horizontally.
%and diagonal up directions, respectively.
On the other hand, Pattern 3 could interpolate feature vectors not only horizontally, but also for viewpoints in diagonal directions. Therefore, we expect that, when taking pictures in a real environment, it will be possible to generate a high-quality volume scene with a few viewpoint images by taking pictures from viewpoints from which feature vectors can be interpolated more variously.
S
between these directions. 

\section{Conclusion}
In this study, we proposed ManifoldNeRF, a new method for few-shot novel view synthesis. On the basis of the assumption that the feature vectors obtained from the pre-trained model also change continuously when the viewpoint of the image changes continuously, the feature vectors at unknown viewpoints nearby known viewpoints can be obtained by interpolation. 

Experimental results show that ManifoldNeRF performs well when the known viewpoints are uniformly selected for the scene. In addition, we clarified which viewpoint pattern is better in real environments, and established a basic policy for practical applications.

In future work, we aim to improve performance further and apply the proposed NeRF method to real-world applications using robots. 

In this study, we considered that optimization for various unknown viewpoints is important for improving performance, so we employed uniform sampling. Although not addressed in this study, a strategy that focuses sampling near the centre of the two known viewpoints is based on the assumption that the area around the known viewpoints can be trained without using pseudo ground truth. From the above, we believe that the accuracy can be further improved by changing the method of selecting unknown viewpoints.

In real-world environments, robots may encounter unknown objects, which they need to learn by themselves. A large number of object images are required for training, but it is time-consuming for the robot to take a large number of images by its own observation. Using the proposed method, we can obtain a volumetric representation of an unknown object from a few images. From this volumetric representation, we can generate a large data set, which can be used for object recognition and other purposes.

\section*{Acknowledgement}
This work was supported by JSPS KAKENHI Grant Number JP21H03519.

\bibliographystyle{unsrt}
\bibliography{bibdata}

% Supplementary material
\setcounter{section}{0}
\renewcommand{\thesection}{\Alph{section}}

\section{Experimental details}
The proposed method was trained by following a training process shown in Algorithm~\ref{alg:manifoldnerf}. The hyperparameters in the training process, manifold loss interval $K$ and scaling factor $\lambda$, were set to 10 and 0.1, respectively. These are the same values as used in the official implementation of DietNeRF~\cite{Jain2021-ru}.

\begin{algorithm}[tb]
\footnotesize
\SetAlgoLined
\KwData{Known viewpoints $\mathcal{D}=\{(I, \mathbf{p})\}$, a pre-trained feature extractor $\phi(\cdot)$, \\ threshold of distance between viewpoints $\epsilon$, manifold loss interval $K$,
scaling factor $\lambda$, batch size $|\mathcal{R}|$, learning rate $\eta_i$, MSE loss $\mathcal{L}_{\text{MSE}}$, manifold loss $\mathcal{L}_{\text{ML}}$}
\KwResult{Trained Neural Radiance Field $f_\theta(\cdot, \cdot)$}
 Initialize NeRF $f_\theta(\cdot, \cdot)$\;
 Pre-compute feature vectors $\mathcal{V} = \{\mathbf{v}=\phi(I) : (I, \mathbf{p}) \in \mathcal{D}\}$\;
 Pre-compute pairs of viewpoint $\mathcal{P} =\{(\{\mathbf{p}_{k,1}, \mathbf{v}_{k,1}\},\{\mathbf{p}_{k,2}, \mathbf{v}_{k,2}\}): (I_{k,1}, \mathbf{p}_{k,1}), (I_{k,2}, \mathbf{p}_{k,2}) \in \mathcal{D}, \mathbf{v}_{k,1}, \mathbf{v}_{k,2} \in \mathcal{V}
 \;, if \; |\mathbf{p}_{k,1} - \mathbf{p}_{k,2}| < \epsilon \}$ \;
 \For{i from 1 to num\_iters}{
  Sample ray batch $\mathcal{R}$, ground-truth colors $\mathbf{C}(\cdot)$\;
  Render rays $\widehat{\mathbf{C}}(\cdot)$\;
  $\mathcal{L} \leftarrow \mathcal{L}_{\text{MSE}}(\mathcal{R}, \mathbf{C}, \widehat{\mathbf{C}})$\;
  \If{$i~\%~K = 0$}{
      Sample pair of viewpoints $(\{\mathbf{p}_k^1, \mathbf{v}_{k,1}\}, \{\mathbf{p}_{k,2}, \mathbf{v}_{k,2}\}) \sim\mathcal{P}$\;
      Compute interpolation coefficient $s$\;
      Compute unknown viewpoint $\widehat{\mathbf{p}}_u = SLERP(\mathbf{p}_{k,1}, \mathbf{p}_{k,2}, s)$\;
      Render image $\widehat{I}$ at viewpoint $\widehat{\mathbf{p}}_u$\;
      Compute feature vector of $\widehat{I}$: $\widehat{\mathbf{v}}_u = \phi{(\widehat{I})}$\;
      Interpolate feature vector $\mathbf{v}_u = LERP(\mathbf{v}_{k,1}, \mathbf{v}_{k,2}, s)$
      
      $\mathcal{L} \leftarrow \mathcal{L} + \mathcal{L}_{\text{ML}}(\mathbf{v}_u, \widehat{\mathbf{v}}_u)$\;
     }
    Update parameters: $\theta \leftarrow Adam(\theta, \eta_i, \nabla_\theta \mathcal{L})$\;
 }
 \caption{Training process of ManifoldNeRF}
 \label{alg:manifoldnerf}
\end{algorithm}

The IDs of the known viewpoints used in the randomly selected experiments from the NeRF synthetic dataset~\cite{Mildenhall2020-ma} were $[2, 16, 26, 55, 73, 75, 86, 93]$, and those used in the DTU MVS dataset were $[0, 6, 7, 23, 32, 37, 39, 48]$.

\section{Analysis of feature vector obtained from pre-trained feature extractor}
In this section, we describe experiments conducted to verify the changes in feature vectors with a change in viewpoints.
For this experiment, we generated 36 images rendered by rotating the camera position by 10 degrees around an axis of the LEGO scene in the NeRF synthetic dataset.
We input these images to the vision encoder of the CLIP to obtain the feature vectors.
The obtained feature vectors were projected onto a 2D space using UMAP and visualized in the 2D space to confirm the changes in feature vectors along with continuous changes in viewpoints.

\begin{figure}[tb]
    \centering
    \includegraphics[width=\linewidth]{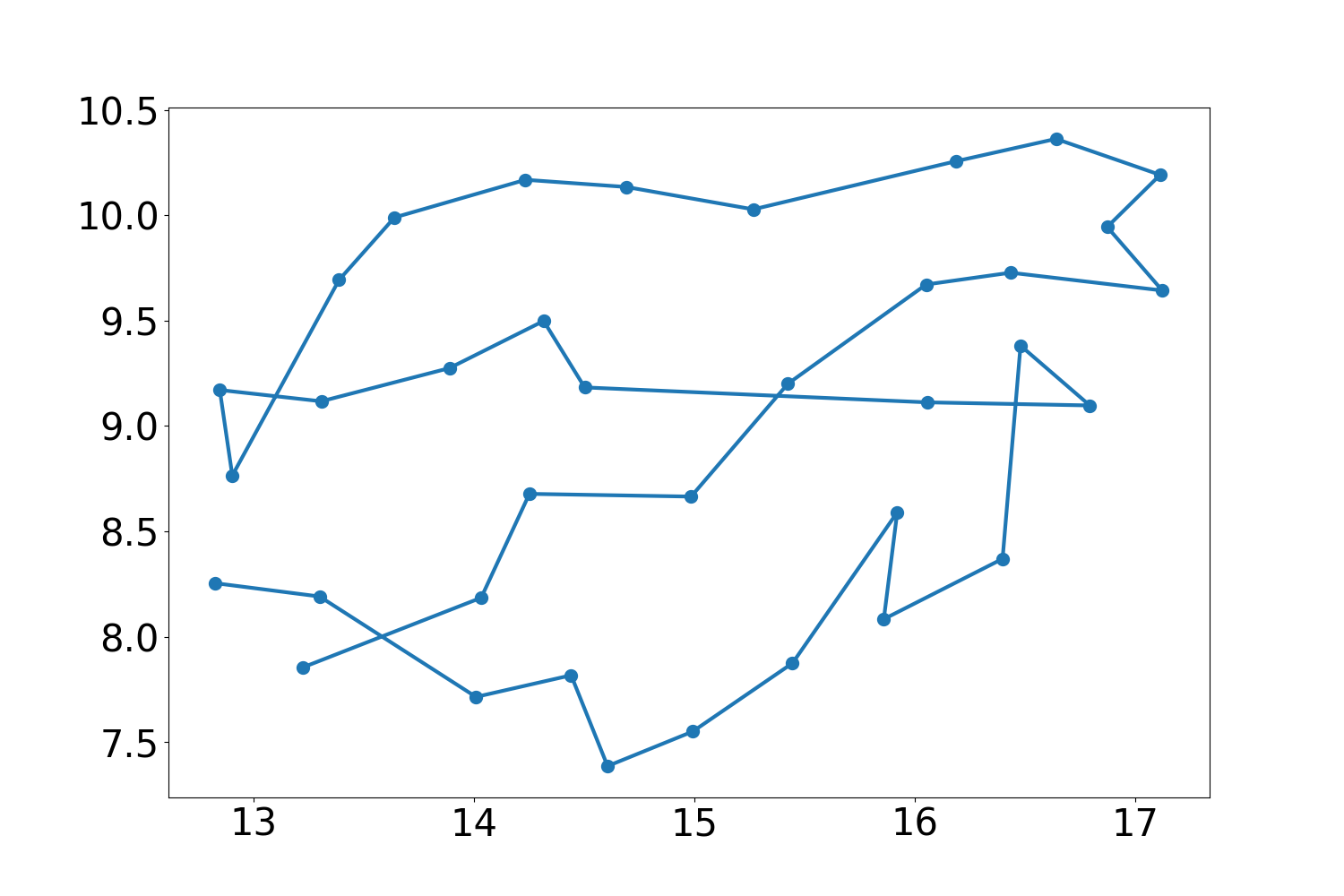}
    \caption{Projected feature vectors extracted with the pre-trained feature extractor into a two-dimensional space using UMAP.
             The dots in the graph denote the projected feature vectors, and the lines connect the dots corresponding to adjacent viewpoints.}
    \label{fig:lego_umap_compressed}
\end{figure}

The experimental results are shown in Fig.~\ref{fig:lego_umap_compressed}.
The feature vectors of adjacent viewpoints are located in the neighborhood, indicating that the feature vectors change continuously as the viewpoints change, as claimed by the Parametric Eigenspace~\cite{Murase1995-bu}.

\section{Performance with different numbers of training data}
In the experiments conducted in Sec. 3 of the paper, the feature vectors between viewpoints differing by 90 degrees were calculated to be close to the ground truth. Therefore, we assumed that the performance of the proposed method would be higher when we selected 8 images if we prepared viewpoints that differ by 90 degrees from each other in the horizontal and the diagonal viewpoints. 

In this section, we evaluated the change in performance when the number of training data is changed. Table~\ref{tab:change_the_number_of_training_data} shows the experimental results when we changed the training data for NeRF and ManifoldNeRF to 4, 8, 12, and 16. NeRF perform poorly when the training data is less than 16. However, the performance of ManifoldNeRF increased significantly when the training data was 8 and above.

\begin{table}[tb]
\caption{Results of performance change with different number of training data. The training dataset is a LEGO scene from the NeRF synthetic dataset.}
\label{tab:change_the_number_of_training_data}
    \begin{minipage}[b]{0.48\columnwidth}
        \centering
        \scalebox{0.75}{
            \begin{tabular}{@{}lccccc@{}}
            \toprule
            \textbf{NeRF} & \textbf{PSNR} $\uparrow$ & \textbf{SSIM} $\uparrow$ & \textbf{LPIPS} $\downarrow$ \\ \midrule
            4   & 9.035  & 0.504 & 0.468 \\
            8   & 9.727  & 0.521 & 0.469 \\
            12  & 10.001 & 0.552 & 0.455 \\
            16  & 27.613 & 0.926 & 0.065 \\\bottomrule
            \end{tabular}
        }
    \end{minipage}
    \hspace{0.01\columnwidth}
    \begin{minipage}[b]{0.48\columnwidth}
        \centering
        \scalebox{0.75}{
            \begin{tabular}{@{}lccccc@{}}
            \toprule
            \textbf{ManifoldNeRF} & \textbf{PSNR} $\uparrow$ & \textbf{SSIM} $\uparrow$ & \textbf{LPIPS} $\downarrow$ \\ \midrule
            4   & 9.219   & 0.526 & 0.464 \\
            8   & 21.607  & 0.806 & 0.174 \\
            12  & 25.215  & 0.874 & 0.108 \\
            16  & 25.793  & 0.884 & 0.102 \\\bottomrule
            \end{tabular}
        }
    \end{minipage}
\end{table}

\section{Details of experimental results using the DTU MVS dataset}
\label{ex:dtu_all}
\begin{figure}[tb]
\centering
    \begin{minipage}[t]{0.18\columnwidth}
        \centering
        \includegraphics[width=\columnwidth]{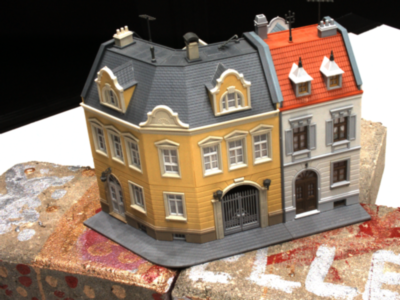}
        \includegraphics[width=\columnwidth]{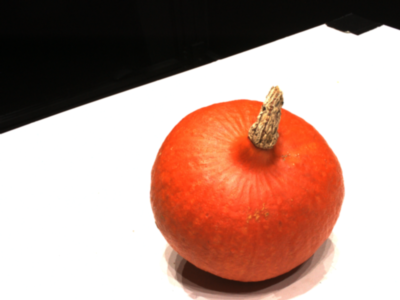}
        \includegraphics[width=\columnwidth]{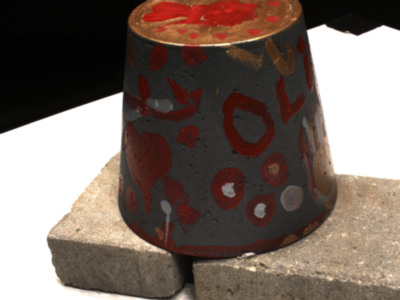}
        \includegraphics[width=\columnwidth]{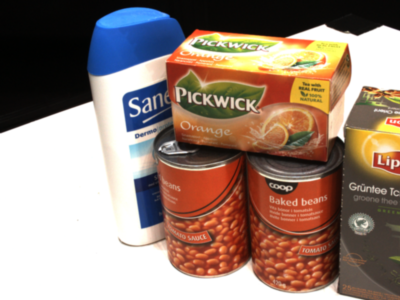}
        \includegraphics[width=\columnwidth]{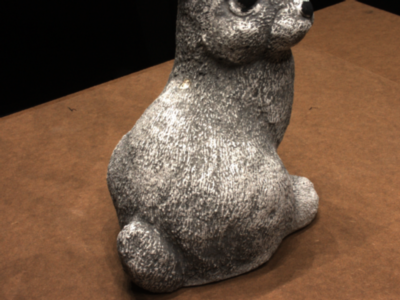}
        \includegraphics[width=\columnwidth]{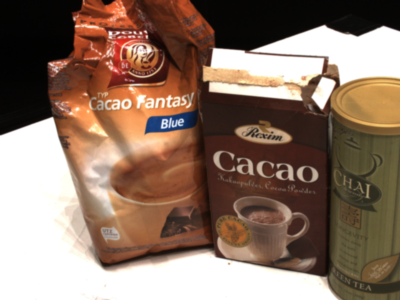}
        \includegraphics[width=\columnwidth]{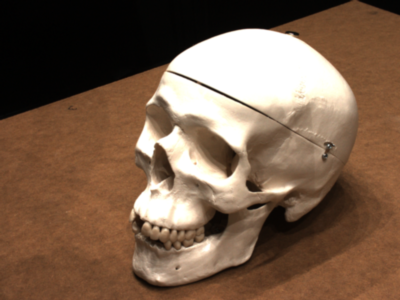}
        \includegraphics[width=\columnwidth]{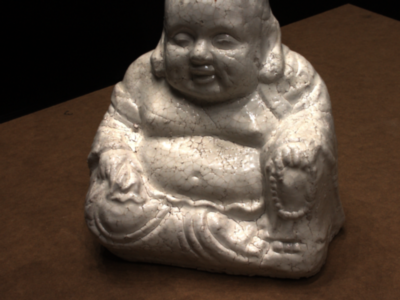}
        Ground Truth  
    \end{minipage}
    \begin{minipage}[t]{0.18\columnwidth}
        \centering
        \includegraphics[width=\columnwidth]{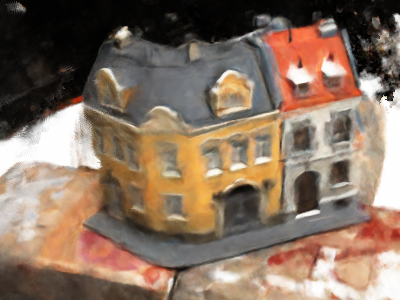}
        \includegraphics[width=\columnwidth]{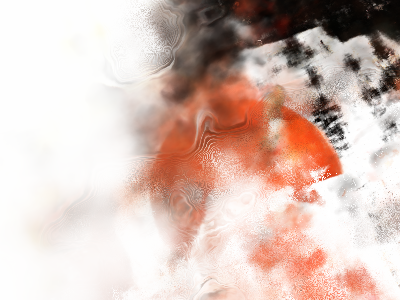}
        \includegraphics[width=\columnwidth]{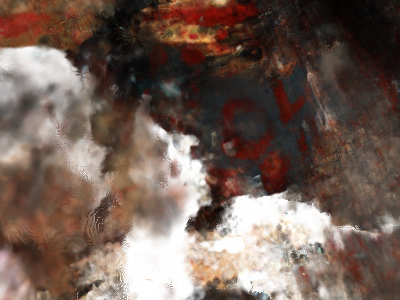}
        \includegraphics[width=\columnwidth]{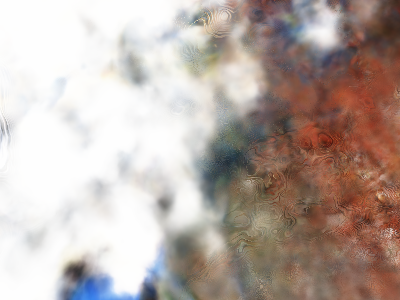}
        \includegraphics[width=\columnwidth]{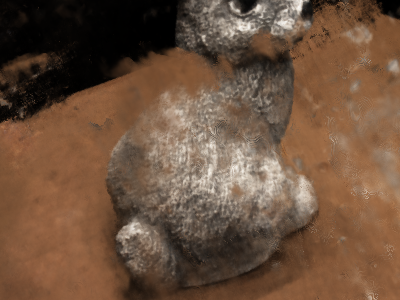}
        \includegraphics[width=\columnwidth]{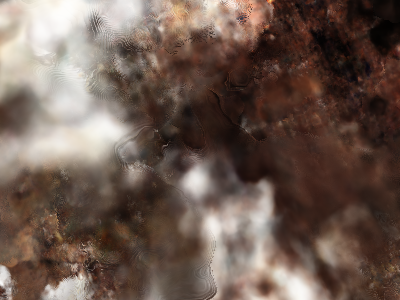}
        \includegraphics[width=\columnwidth]{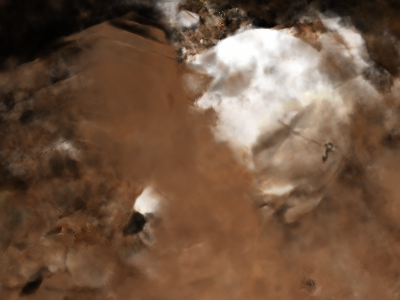}
        \includegraphics[width=\columnwidth]{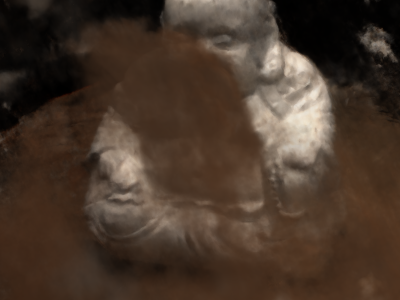}
        NeRF  
    \end{minipage}
    \begin{minipage}[t]{0.18\columnwidth}
        \centering
        \includegraphics[width=\columnwidth]{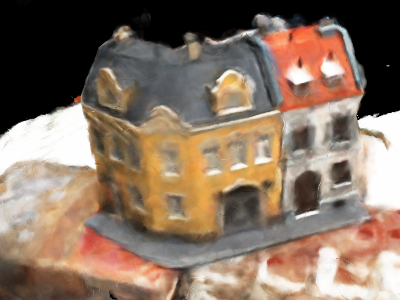}
        \includegraphics[width=\columnwidth]{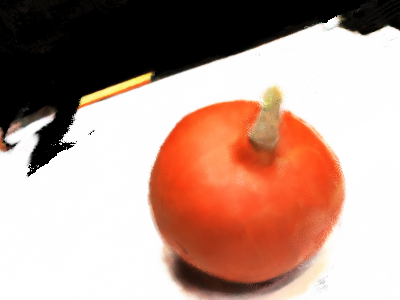}
        \includegraphics[width=\columnwidth]{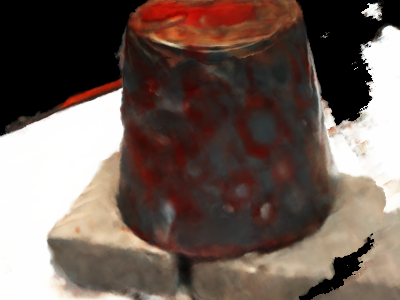}
        \includegraphics[width=\columnwidth]{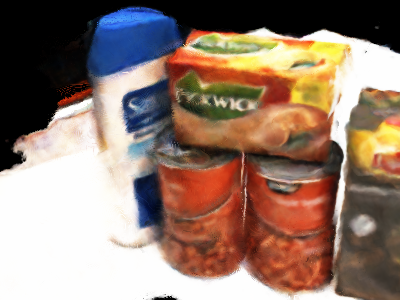}
        \includegraphics[width=\columnwidth]{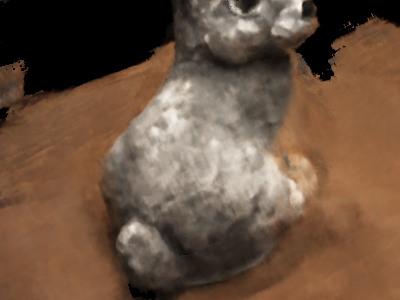}
        \includegraphics[width=\columnwidth]{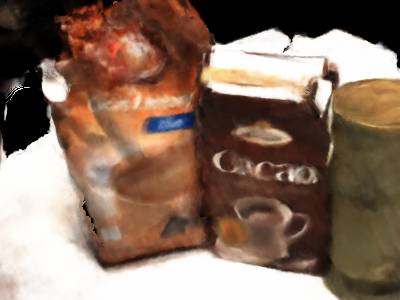}
        \includegraphics[width=\columnwidth]{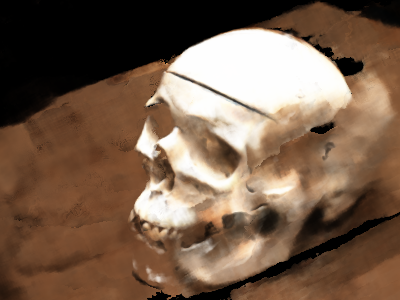}
        \includegraphics[width=\columnwidth]{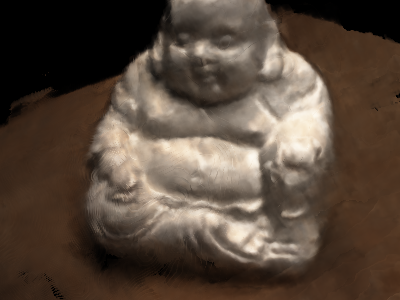}
        InfoNeRF  
    \end{minipage}
    \begin{minipage}[t]{0.18\columnwidth}
        \centering
        \includegraphics[width=\columnwidth]{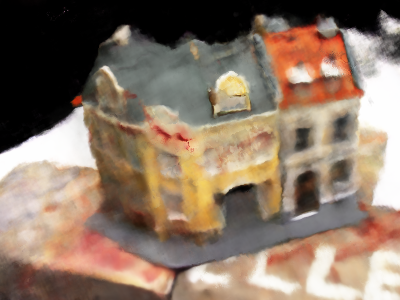}
        \includegraphics[width=\columnwidth]{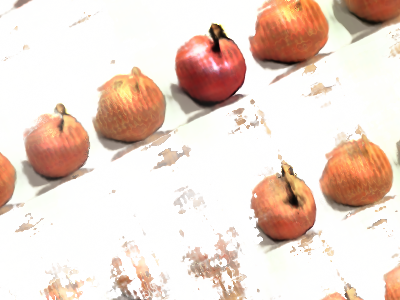}
        \includegraphics[width=\columnwidth]{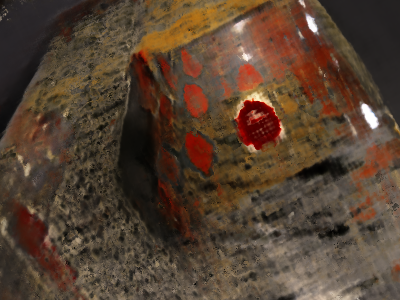}
        \includegraphics[width=\columnwidth]{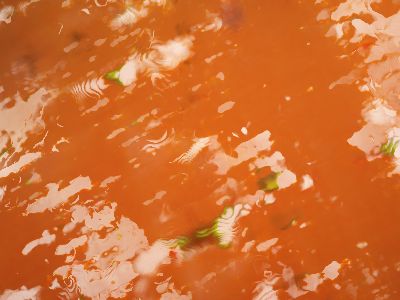}
        \includegraphics[width=\columnwidth]{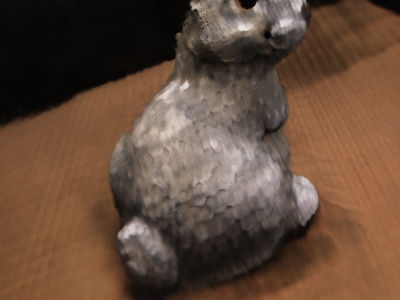}
        \includegraphics[width=\columnwidth]{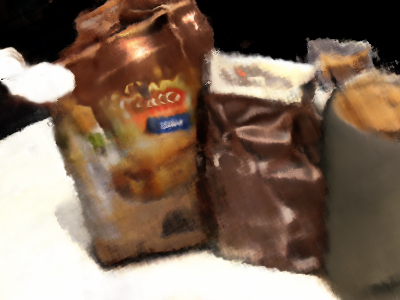}
        \includegraphics[width=\columnwidth]{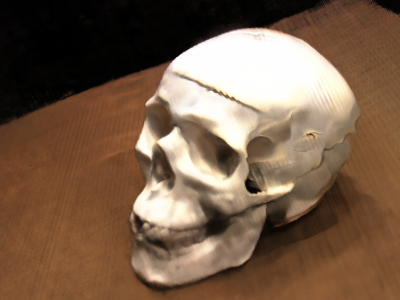}
        \includegraphics[width=\columnwidth]{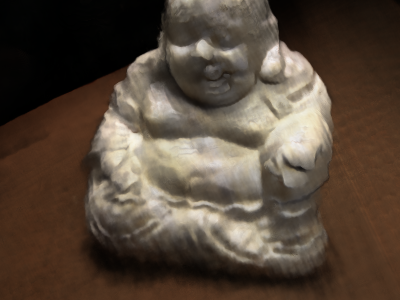}
        DietNeRF
    \end{minipage}
    \begin{minipage}[t]{0.18\columnwidth}
        \centering
        \includegraphics[width=\columnwidth]{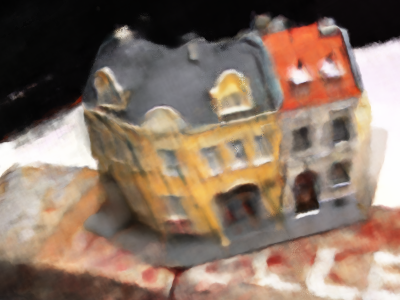}
        \includegraphics[width=\columnwidth]{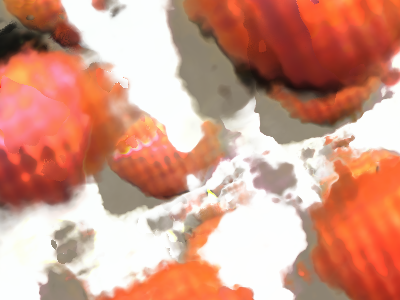}
        \includegraphics[width=\columnwidth]{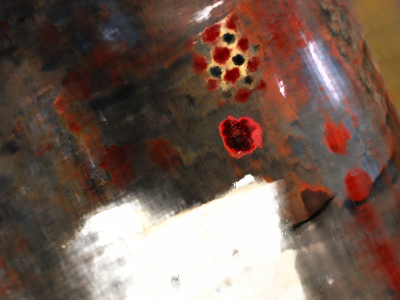}
        \includegraphics[width=\columnwidth]{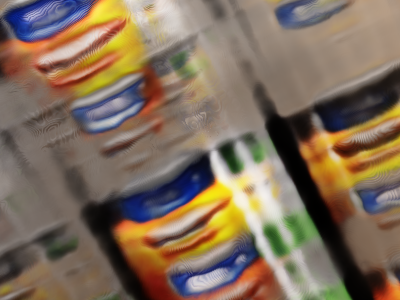}
        \includegraphics[width=\columnwidth]{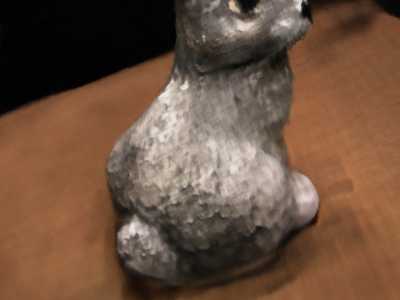}
        \includegraphics[width=\columnwidth]{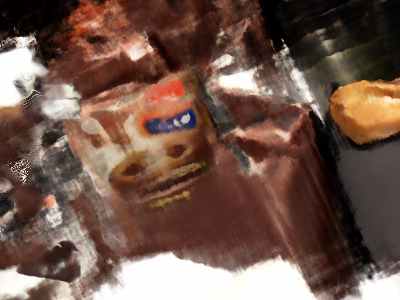}
        \includegraphics[width=\columnwidth]{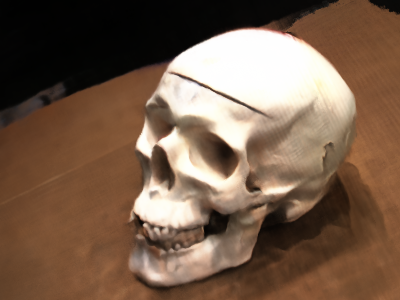}
        \includegraphics[width=\columnwidth]{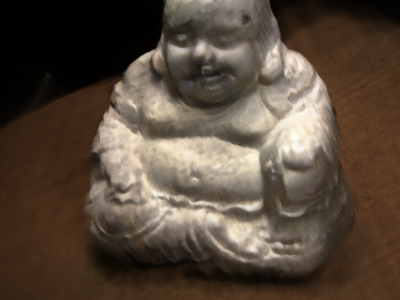}
        ManifoldNeRF (ours)
    \end{minipage}
    \caption{Qualitative comparison on 8 scenes of the MVS DTU dataset}
    \label{fig:dtu_full}
\end{figure}

Table~\ref{tab:dtu_full_result} and Fig.~\ref{fig:dtu_full}  show the experimental results for 8 scenes selected from the DTU MVS dataset~\cite{DTU}.
ManifoldNeRF performed well in 4 scenes of the 8 scenes.
However, in the remaining 4 scenes, the performance was comparable to that of vanilla NeRF.
The reason why the proposed method sometimes performed not well is that the performance of ManiofldNeRF is strongly dependent on the location of known viewpoints.
In contrast, InfoNeRF~\cite{kim2022infonerf} performed more stably than the other methods.

\begin{table}[tb]
\caption{Results of training 8 randomly selected images in 8 scene of the DTU MVS dataset. The highest score is in bold, and the second-highest score is underlined.}
\label{tab:dtu_full_result}
    \begin{minipage}[b]{0.48\columnwidth}
        \centering
        \scalebox{0.75}{
            \begin{tabular}{@{}lccccc@{}}
            \toprule
            \textbf{\#6} & \textbf{PSNR} $\uparrow$ & \textbf{SSIM} $\uparrow$ & \textbf{LPIPS} $\downarrow$ \\ \midrule
            NeRF & \underline{15.550} & \underline{0.460} & 0.472 \\
            InfoNeRF & 13.352 & 0.397 & \underline{0.462}  \\
            DietNeRF & 15.210 & 0.426 & 0.476  \\
            ManifoldNeRF (ours) & \textbf{16.232} & \textbf{0.508} & \textbf{0.451}  \\\bottomrule
            \end{tabular}
        }
    \end{minipage}
    \hspace{0.01\columnwidth}
    \begin{minipage}[b]{0.48\columnwidth}
        \centering
        \scalebox{0.75}{
            \begin{tabular}{@{}lccccc@{}}
            \toprule
            \textbf{\#56} & \textbf{PSNR} $\uparrow$ & \textbf{SSIM} $\uparrow$ & \textbf{LPIPS} $\downarrow$ \\ \midrule
            NeRF & \underline{21.484} & \underline{0.621} & \textbf{0.353} \\
            InfoNeRF & 18.644 & 0.477 & 0.474  \\
            DietNeRF & 19.026 & 0.538 & 0.427  \\
            ManifoldNeRF (ours) & \textbf{22.197} & \textbf{0.639} & \underline{0.367}  \\\bottomrule
            \end{tabular}
        }
    \end{minipage}

    \begin{minipage}[b]{0.48\columnwidth}
        \centering
        \scalebox{0.75}{
            \begin{tabular}{@{}lccccc@{}}
            \toprule
            \textbf{\#65} & \textbf{PSNR} $\uparrow$ & \textbf{SSIM} $\uparrow$ & \textbf{LPIPS} $\downarrow$ \\ \midrule
            NeRF & 11.970 & 0.481 & 0.527 \\
            InfoNeRF & 14.786 & 0.484 & 0.431  \\
            DietNeRF & \underline{20.883} & \underline{0.698} & \underline{0.352}  \\
            ManifoldNeRF (ours) & \textbf{22.197} & \textbf{0.702} & \textbf{0.302}  \\\bottomrule
            \end{tabular}
        }
    \end{minipage}
    \hspace{0.01\columnwidth}
    \begin{minipage}[b]{0.48\columnwidth}
        \centering
        \scalebox{0.75}{
            \begin{tabular}{@{}lccccc@{}}
            \toprule
            \textbf{\#114}         & \textbf{PSNR} $\uparrow$ & \textbf{SSIM} $\uparrow$ & \textbf{LPIPS} $\downarrow$ \\ \midrule
            NeRF & 18.691 & 0.636 & 0.396  \\
            InfoNeRF & \underline{21.382} & 0.611 & 0.364  \\
            DietNeRF & 20.861 & \underline{0.673} & \underline{0.337}  \\
            ManifoldNeRF (ours) & \textbf{23.202} & \textbf{0.732} & \textbf{0.299}  \\\bottomrule
            \end{tabular}
        }
    \end{minipage}
    
    \begin{minipage}[b]{0.48\columnwidth}
        \centering
        \scalebox{0.75}{
            \begin{tabular}{@{}lccccc@{}}
            \toprule
            \textbf{\#30} & \textbf{PSNR} $\uparrow$ & \textbf{SSIM} $\uparrow$ & \textbf{LPIPS} $\downarrow$ \\ \midrule
            NeRF & \underline{8.054} & \underline{0.491} & \underline{0.560} \\
            InfoNeRF & \textbf{17.657} & \textbf{0.663} & \textbf{0.254}  \\
            DietNeRF & 6.092 & 0.298 & 0.675  \\
            ManifoldNeRF (ours) & 6.406 & 0.387 & 0.633  \\\bottomrule
            \end{tabular}
        }
    \end{minipage}
    \hspace{0.01\columnwidth}
    \begin{minipage}[b]{0.48\columnwidth}
        \centering
        \scalebox{0.75}{
            \begin{tabular}{@{}lccccc@{}}
            \toprule
            \textbf{\#41} & \textbf{PSNR} $\uparrow$ & \textbf{SSIM} $\uparrow$ & \textbf{LPIPS} $\downarrow$ \\ \midrule
            NeRF & 8.236 & 0.312 & 0.636 \\
            InfoNeRF & \textbf{14.681} & \textbf{0.484} & \textbf{0.423}  \\
            DietNeRF & 8.36 & 0.246 & 0.636  \\
            ManifoldNeRF (ours) & \underline{8.963} & \underline{0.317} & \underline{0.628}  \\\bottomrule
            \end{tabular}
        }
    \end{minipage}
    
    \begin{minipage}[b]{0.48\columnwidth}
        \centering
        \scalebox{0.75}{
            \begin{tabular}{@{}lccccc@{}}
            \toprule
            \textbf{\#45} & \textbf{PSNR} $\uparrow$ & \textbf{SSIM} $\uparrow$ & \textbf{LPIPS} $\downarrow$ \\ \midrule
            NeRF & \underline{7.558} & \underline{0.220} & 0.710 \\
            InfoNeRF & \textbf{10.719} & \textbf{0.422} & \textbf{0.441}  \\
            DietNeRF & 7.097 & 0.216 & \underline{0.662}  \\
            ManifoldNeRF (ours) & 7.418 & 0.181 & 0.721  \\\bottomrule
            \end{tabular}
        }
    \end{minipage}
    \hspace{0.01\columnwidth}
    \begin{minipage}[b]{0.48\columnwidth}
        \centering
        \scalebox{0.75}{
            \begin{tabular}{@{}lccccc@{}}
            \toprule
            \textbf{\#61} & \textbf{PSNR} $\uparrow$ & \textbf{SSIM} $\uparrow$ & \textbf{LPIPS} $\downarrow$ \\ \midrule
            NeRF & 7.793 & 0.236 & 0.684 \\
            InfoNeRF & \textbf{14.634} & \textbf{0.543} & \textbf{0.395}  \\
            DietNeRF & \underline{11.974} & \underline{0.463} & \underline{0.508}  \\
            ManifoldNeRF (ours) & 7.518 & 0.267 & 0.679 \\\bottomrule
            \end{tabular}
        }
    \end{minipage}

\end{table}

\begin{figure}[tb]
\centering
    \begin{minipage}[t]{0.14\columnwidth}
        \centering
        \includegraphics[width=\columnwidth]{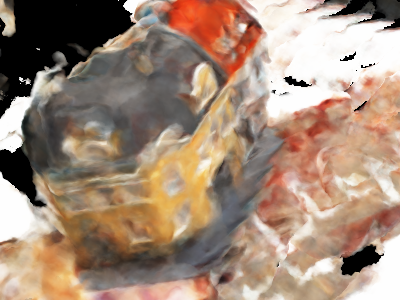}
        \includegraphics[width=\columnwidth]{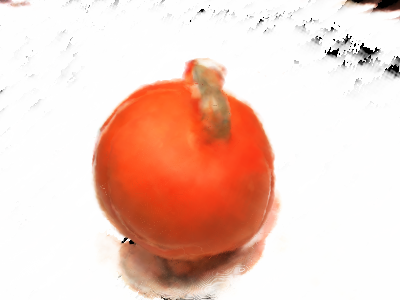}
        \includegraphics[width=\columnwidth]{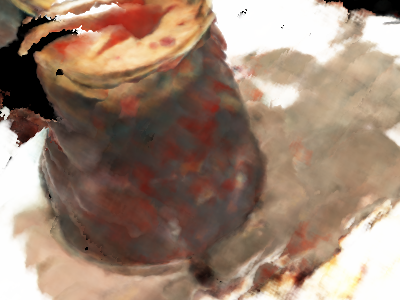}
        \includegraphics[width=\columnwidth]{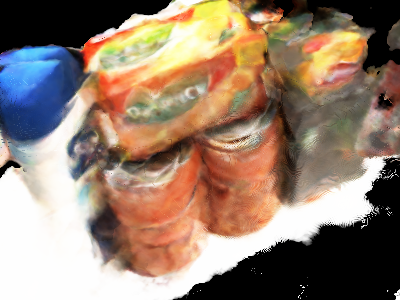}
        \includegraphics[width=\columnwidth]{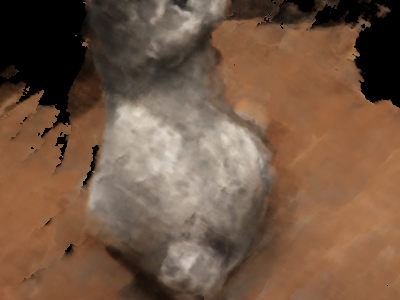}
        \includegraphics[width=\columnwidth]{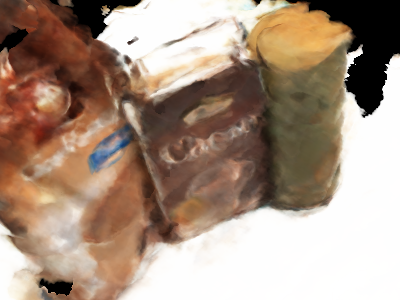}
        \includegraphics[width=\columnwidth]{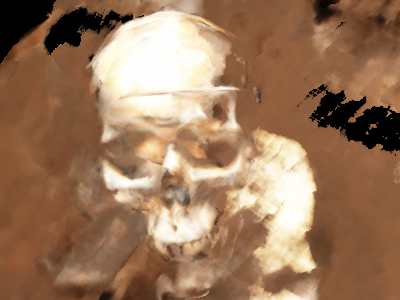}
        \includegraphics[width=\columnwidth]{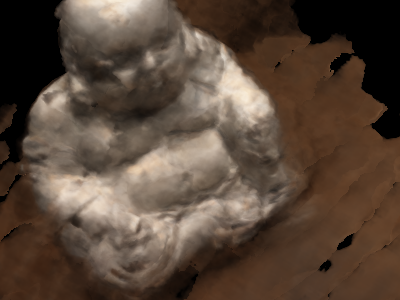}
    \end{minipage}
    \begin{minipage}[t]{0.14\columnwidth}
        \centering
        \includegraphics[width=\columnwidth]{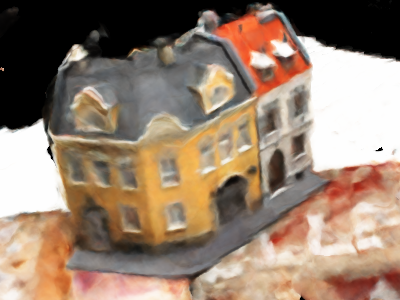}
        \includegraphics[width=\columnwidth]{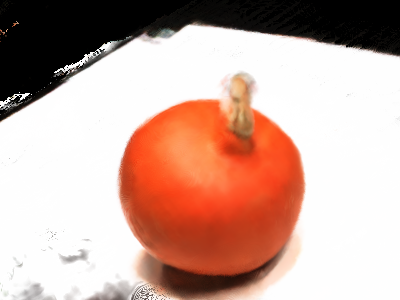}
        \includegraphics[width=\columnwidth]{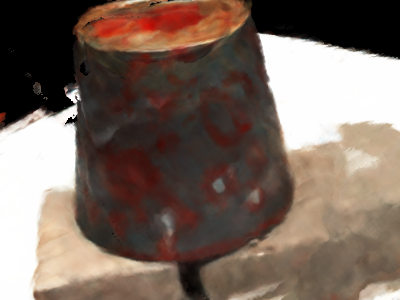}
        \includegraphics[width=\columnwidth]{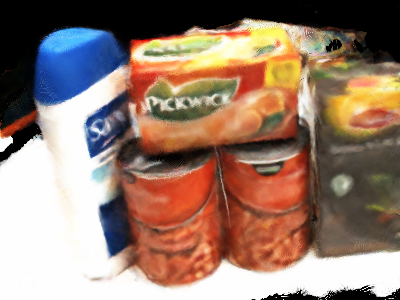}
        \includegraphics[width=\columnwidth]{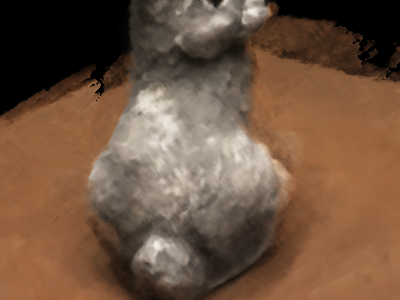}
        \includegraphics[width=\columnwidth]{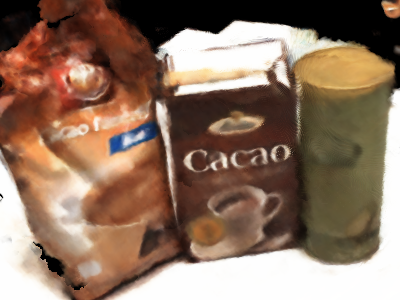}
        \includegraphics[width=\columnwidth]{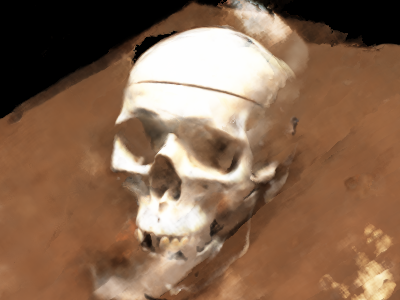}
        \includegraphics[width=\columnwidth]{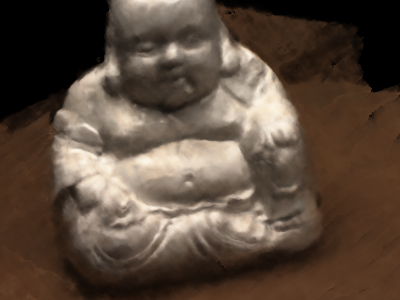}
        w/o fine-tuning
    \end{minipage}
    \begin{minipage}[t]{0.14\columnwidth}
        \centering
        \includegraphics[width=\columnwidth]{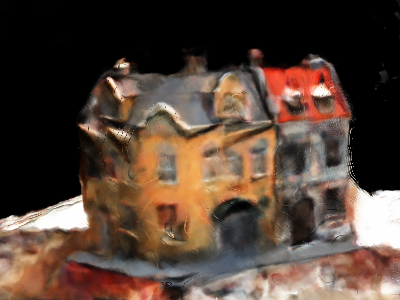}
        \includegraphics[width=\columnwidth]{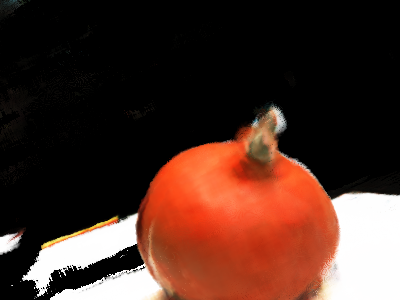}
        \includegraphics[width=\columnwidth]{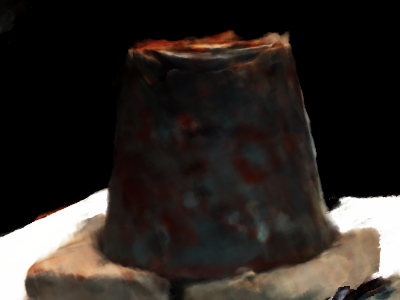}
        \includegraphics[width=\columnwidth]{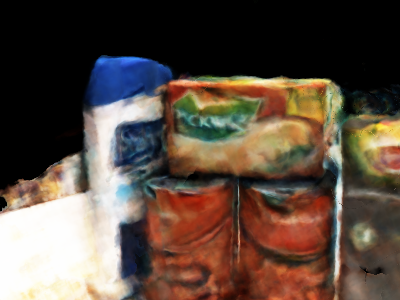}
        \includegraphics[width=\columnwidth]{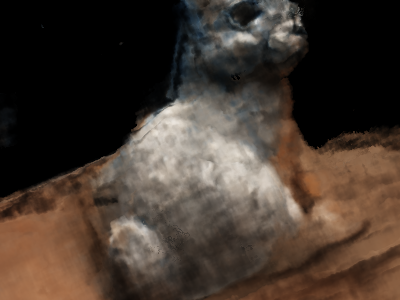}
        \includegraphics[width=\columnwidth]{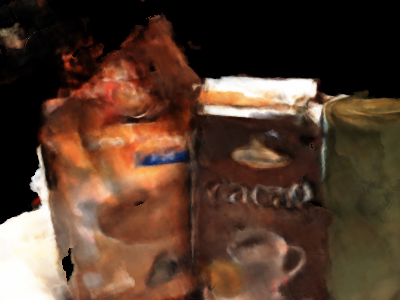}
        \includegraphics[width=\columnwidth]{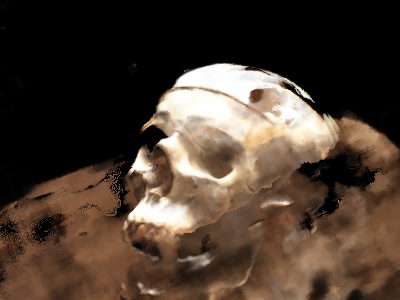}
        \includegraphics[width=\columnwidth]{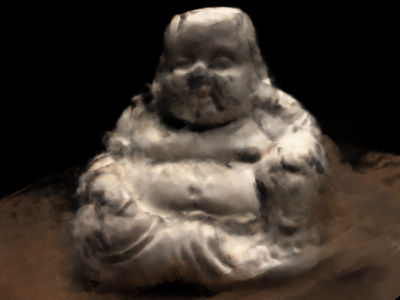}
    \end{minipage}
    \begin{minipage}[t]{0.15\columnwidth}
        
    \end{minipage}
    \begin{minipage}[t]{0.14\columnwidth}
        \centering
        \includegraphics[width=\columnwidth]{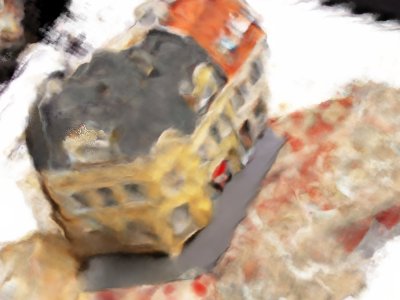}
        \includegraphics[width=\columnwidth]{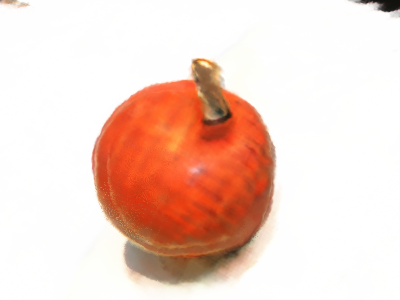}
        \includegraphics[width=\columnwidth]{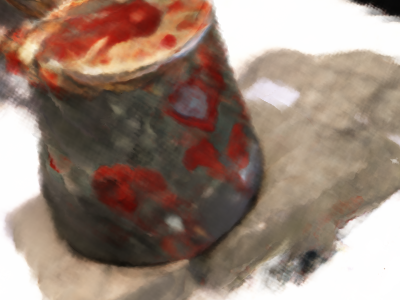}
        \includegraphics[width=\columnwidth]{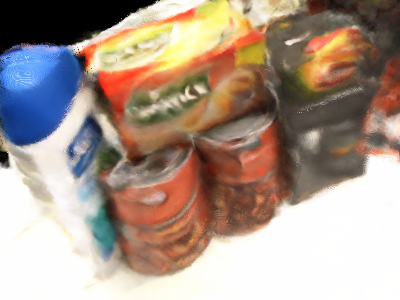}
        \includegraphics[width=\columnwidth]{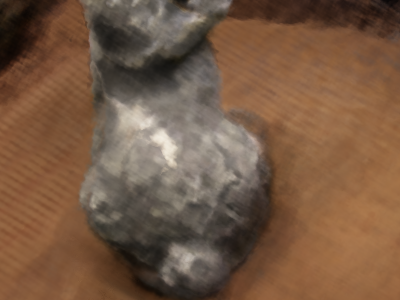}
        \includegraphics[width=\columnwidth]{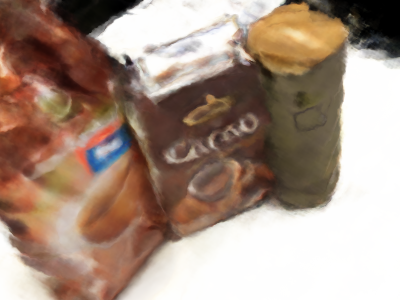}
        \includegraphics[width=\columnwidth]{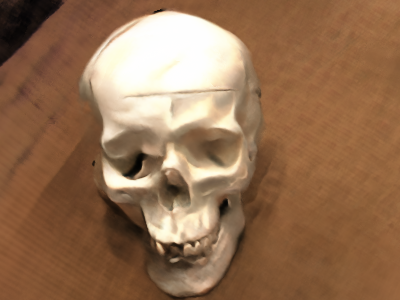}
        \includegraphics[width=\columnwidth]{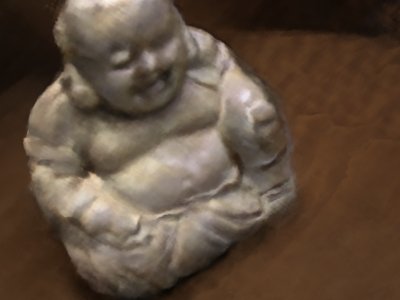}
    \end{minipage}
    \begin{minipage}[t]{0.14\columnwidth}
        \centering
        \includegraphics[width=\columnwidth]{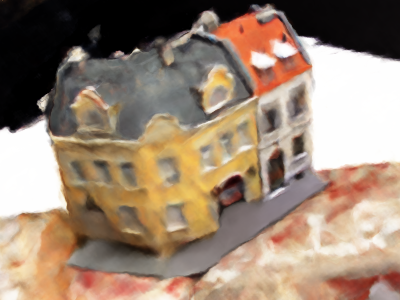}
        \includegraphics[width=\columnwidth]{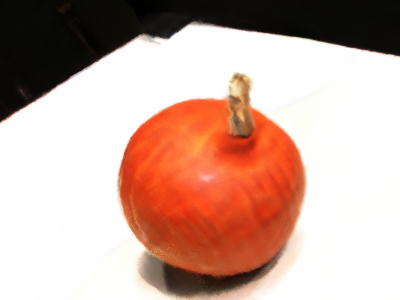}
        \includegraphics[width=\columnwidth]{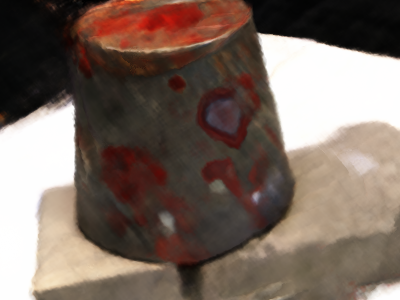}
        \includegraphics[width=\columnwidth]{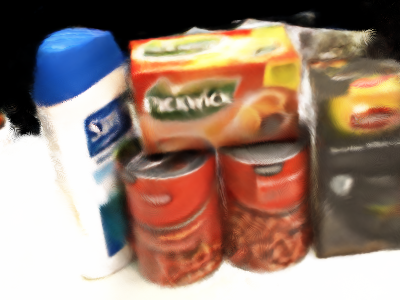}
        \includegraphics[width=\columnwidth]{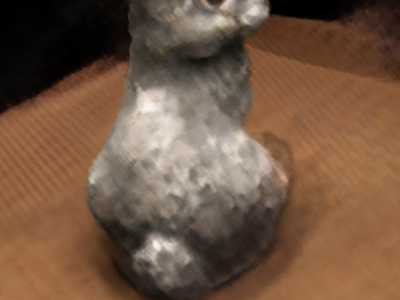}
        \includegraphics[width=\columnwidth]{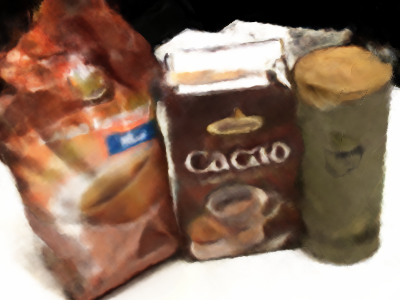}
        \includegraphics[width=\columnwidth]{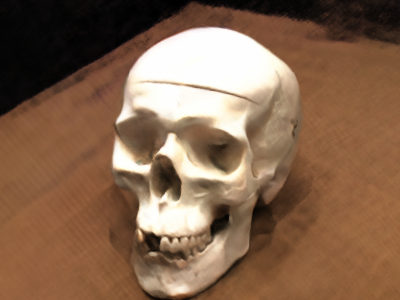}
        \includegraphics[width=\columnwidth]{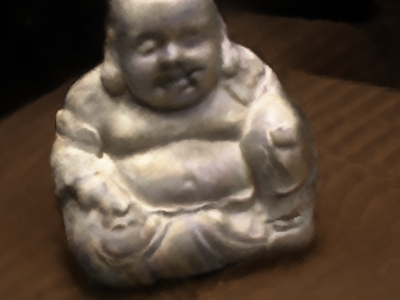}
        w/ fine-tuning 100k iters
    \end{minipage}
    \begin{minipage}[t]{0.14\columnwidth}
        \centering
        \includegraphics[width=\columnwidth]{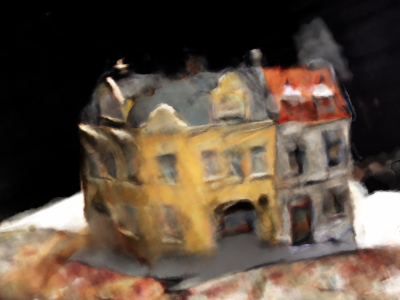}
        \includegraphics[width=\columnwidth]{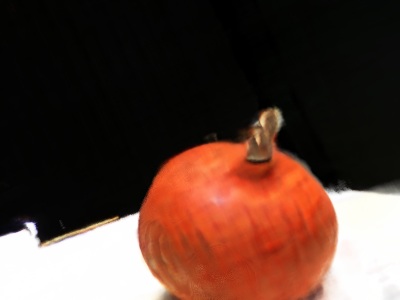}
        \includegraphics[width=\columnwidth]{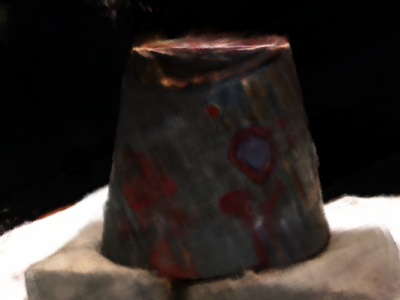}
        \includegraphics[width=\columnwidth]{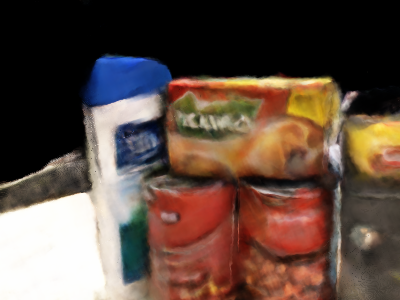}
        \includegraphics[width=\columnwidth]{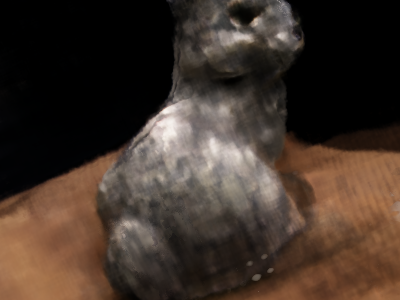}
        \includegraphics[width=\columnwidth]{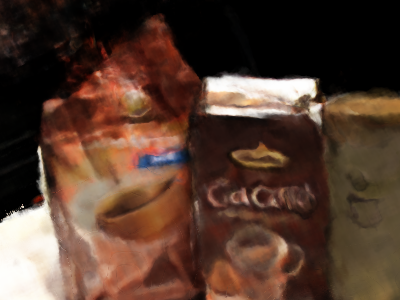}
        \includegraphics[width=\columnwidth]{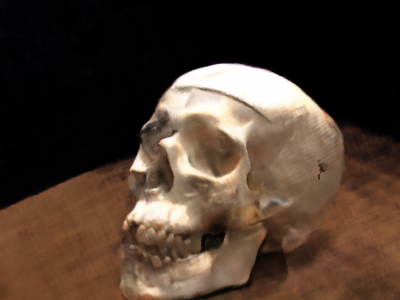}
        \includegraphics[width=\columnwidth]{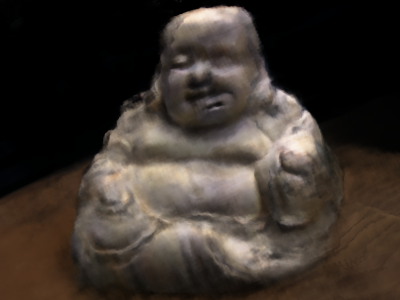}
    \end{minipage}
    \caption{Qualitative results of fine-tuning.}
    \label{fig:dtu_ft}
\end{figure}

Next, Table~\ref{tab:dtu_finetune} and Fig.~\ref{fig:dtu_ft} show the results of fine-tuning the model trained with InfoNeRF using ManifoldNeRF.
We confirmed the performance improvement by fine-tuning the model trained with InfoNeRF.
The reason for the performance improvement is that InfoNeRF applies constraints to each point on the ray, whereas ManifoldNeRF applies constraints to the feature vectors of the image obtained from the viewpoints between neighbouring viewpoints, and the optimization targets are different.
From the above, we demonstrate that even when the known viewpoints are random, combining other methods with the proposed method can improve performance.

\begin{table}[tb]
\caption{Results of fine-tuning the model trained with InfoNeRF using ManifoldNeRF. The results w/o fine-tuning are the same as for InfoNeRF in Table~\ref{tab:dtu_full_result}. The highest score is in bold, and the second-highest score is underlined.}
\label{tab:dtu_finetune}
    \begin{minipage}[b]{0.48\columnwidth}
        \centering
        \scalebox{0.75}{
            \begin{tabular}{@{}lccccc@{}}
            \toprule
            \textbf{\#6} & \textbf{PSNR} $\uparrow$ & \textbf{SSIM} $\uparrow$ & \textbf{LPIPS} $\downarrow$ \\ \midrule
            w/o fine-tuning & 13.352 & 0.397 & 0.462  \\ \hdashline
            20k iters  & 15.157 & 0.438 & 0.491 \\
            40k iters  & 15.210 & 0.448 & 0.476 \\
            60k iters  & 15.087 & 0.461 & \underline{0.458}s \\
            80k iters  & \underline{15.211} & \underline{0.471} & \underline{0.458} \\
            100k iters & \textbf{15.398} & \textbf{0.472} & \textbf{0.448} \\\bottomrule
            \end{tabular}
        }
    \end{minipage}
    \hspace{0.01\columnwidth}
    \begin{minipage}[b]{0.48\columnwidth}
        \centering
        \scalebox{0.75}{
            \begin{tabular}{@{}lccccc@{}}
            \toprule
            \textbf{\#30} & \textbf{PSNR} $\uparrow$ & \textbf{SSIM} $\uparrow$ & \textbf{LPIPS} $\downarrow$ \\ \midrule
            w/o fine-tuning & 17.657 & 0.663 & 0.254  \\ \hdashline
            20k iters  & 20.335 & 0.808 & 0.197 \\
            40k iters  & \textbf{20.412} & 0.828 & 0.185 \\
            60k iters  & \underline{20.364} & \textbf{0.839} & 0.183 \\
            80k iters  & 20.359 & 0.835 & \underline{0.180} \\
            100k iters & 20.285 & \underline{0.837} & \textbf{0.179} \\\bottomrule
            \end{tabular}
        }
    \end{minipage}

    \begin{minipage}[b]{0.48\columnwidth}
        \centering
        \scalebox{0.75}{
            \begin{tabular}{@{}lccccc@{}}
            \toprule
            \textbf{\#41} & \textbf{PSNR} $\uparrow$ & \textbf{SSIM} $\uparrow$ & \textbf{LPIPS} $\downarrow$ \\ \midrule
            w/o fine-tuning & 14.681 & 0.484 & 0.423  \\ \hdashline
            20k iters  & \underline{17.032} & 0.570 & 0.443 \\
            40k iters  & \textbf{17.111} & 0.585 & 0.427  \\
            60k iters  & 16.890 & 0.582 & 0.422 \\
            80k iters  & 16.904 & \underline{0.595} & \underline{0.411} \\
            100k iters & 17.031 & \textbf{0.599} & \textbf{0.405} \\\bottomrule
            \end{tabular}
        }
    \end{minipage}
    \hspace{0.01\columnwidth}
    \begin{minipage}[b]{0.48\columnwidth}
        \centering
        \scalebox{0.75}{
            \begin{tabular}{@{}lccccc@{}}
            \toprule
            \textbf{\#45} & \textbf{PSNR} $\uparrow$ & \textbf{SSIM} $\uparrow$ & \textbf{LPIPS} $\downarrow$ \\ \midrule
            w/o fine-tuning & 10.719 & 0.422 & 0.441  \\ \hdashline
            20k iters  & 14.867 & 0.504 & 0.412 \\
            40k iters  & 14.882 & 0.514 & 0.399 \\
            60k iters  & 14.842 & \underline{0.520} & 0.390  \\
            80k iters  & \underline{14.898} & \textbf{0.523} & \underline{0.389}  \\
            100k iters & \textbf{14.903} & \textbf{0.523} & \textbf{0.383}  \\\bottomrule

            \end{tabular}
        }
    \end{minipage}

    \begin{minipage}[b]{0.48\columnwidth}
        \centering
        \scalebox{0.75}{
            \begin{tabular}{@{}lccccc@{}}
            \toprule
            \textbf{\#56} & \textbf{PSNR} $\uparrow$ & \textbf{SSIM} $\uparrow$ & \textbf{LPIPS} $\downarrow$ \\ \midrule
            w/o fine-tuning & 18.644 & 0.477 & 0.474  \\ \hdashline
            20k iters  & \underline{19.912} & 0.510 & 0.472  \\
            40k iters  & \textbf{20.130} & 0.528 & 0.462 \\
            60k iters  & 19.880 & \underline{0.532} & 0.458  \\
            80k iters  & 19.773 & 0.531 & \underline{0.453}  \\
            100k iters & 19.902 & \textbf{0.540} & \textbf{0.451}  \\\bottomrule
        
            \end{tabular}
        }
    \end{minipage}
    \hspace{0.01\columnwidth}
    \begin{minipage}[b]{0.48\columnwidth}
        \centering
        \scalebox{0.75}{
            \begin{tabular}{@{}lccccc@{}}
            \toprule
            \textbf{\#61} & \textbf{PSNR} $\uparrow$ & \textbf{SSIM} $\uparrow$ & \textbf{LPIPS} $\downarrow$ \\ \midrule
            w/o fine-tuning & 14.634 & 0.543 & 0.395 \\ \hdashline
            20k iters  & 15.588 & 0.544 & 0.437   \\
            40k iters  & \textbf{15.926} & 0.554 & 0.420 \\
            60k iters  & 15.772 & \underline{0.563} & 0.415  \\
            80k iters  & \underline{15.861} & \textbf{0.568} & \underline{0.409}  \\
            100k iters & 15.799 & 0.560 & \textbf{0.403}  \\\bottomrule
            \end{tabular}
        }
    \end{minipage}
    
    \begin{minipage}[b]{0.48\columnwidth}
        \centering
        \scalebox{0.75}{
            \begin{tabular}{@{}lccccc@{}}
            \toprule
            \textbf{\#65} & \textbf{PSNR} $\uparrow$ & \textbf{SSIM} $\uparrow$ & \textbf{LPIPS} $\downarrow$ \\ \midrule
            w/o fine-tuning & 14.786 & 0.484 & 0.431 \\ \hdashline
            20k iters  & 20.468 & 0.658 & 0.369    \\
            40k iters  & 20.615 & 0.673 & 0.349 \\
            60k iters  & 20.654 & 0.687 & 0.339 \\
            80k iters  & \underline{20.768} & \underline{0.693} & \underline{0.334}   \\
            100k iters & \textbf{20.821} & \textbf{0.705} & \textbf{0.331}  \\\bottomrule
            \end{tabular}
        }
    \end{minipage}
    \hspace{0.01\columnwidth}
    \begin{minipage}[b]{0.48\columnwidth}
        \centering
        \scalebox{0.75}{
            \begin{tabular}{@{}lccccc@{}}
            \toprule
            \textbf{\#114}         & \textbf{PSNR} $\uparrow$ & \textbf{SSIM} $\uparrow$ & \textbf{LPIPS} $\downarrow$ \\ \midrule
            w/o fine-tuning & 21.382 & 0.611 & 0.364 \\ \hdashline
            20k iters  & 21.902 & 0.658 & 0.372   \\
            40k iters  & \textbf{22.092} & 0.672 & 0.358  \\
            60k iters  & \underline{21.948} & \textbf{0.677} & \underline{0.350} \\
            80k iters  & 21.791 & 0.673 & \underline{0.350}    \\
            100k iters & 21.696 & \underline{0.675} & \textbf{0.346}  \\\bottomrule
            \end{tabular}
        }
    \end{minipage}
\end{table}

\end{document}